\documentclass{article}

\usepackage{arxiv}

\usepackage{amssymb}
\usepackage{rotating}
\usepackage{floatrow}
\usepackage{multirow}
\usepackage{graphicx}
\usepackage{soul}
\usepackage{wrapfig}
\usepackage{hyperref}

\hypersetup{
    colorlinks=true,
    linkcolor=blue,
    filecolor=blue,      
    urlcolor=blue,
    }

\usepackage{comment}

\usepackage{multicol}
\usepackage{comment}
\usepackage{amsmath}
\usepackage{booktabs, makecell, tabularx}
\usepackage{amsfonts}
\usepackage[dvipsnames,table,xcdraw]{xcolor}
\definecolor{gold}{rgb}{0.85, 0.65, 0.13}
\usepackage{algorithm}
\usepackage[noend]{algpseudocode}

\algnewcommand\algorithmicforeach{\textbf{for each}}
\algdef{S}[FOR]{ForEach}[1]{\algorithmicforeach\ #1\ \algorithmicdo}
\algnewcommand{\algorithmicand}{\textbf{ and }}
\algnewcommand{\algorithmicor}{\textbf{ or }}
\algnewcommand{\OR}{\algorithmicor}
\algnewcommand{\AND}{\algorithmicand}
\algnewcommand{\LineComment}[1]{\State \(\triangleright\) {\footnotesize #1}}
\newcommand{\pluseq}{\mathrel{+}=}
\algdef{SE}[VARIABLES]{Variables}{EndVariables}
   {\algorithmicvariables}
   {\algorithmicend\ \algorithmicvariables}
\algnewcommand{\algorithmicvariables}{\textbf{global variables}}

\usepackage[caption=false]{subfig}
\usepackage{pgfplots}

\pgfplotsset{compat=1.12}

\newcommand{\ie}{{\it i.e.}}

\newcommand{\bplesscants}[0]{\mbox{BP-free~CANTS }}

\title{Backpropagation-Free 4D Continuous Ant-Based Neural Topology Search}

\author{ 
    AbdElRahman ElSaid
	\thanks{Department of Computer Science, University of North Carolina Wilmington}\\
	\texttt{elsaida@uncw.edu} \\
	\AND
	Karl Ricanek~\footnotemark[1] \\
	\texttt{ricanekk@uncw.edu} \\
 	\And
	Zeming Lyu~\thanks{Software Engineering Department, Rochester Institute of Technology} \\
	\texttt{zl7069@rit.edu} \\
    \And
    Alexander Ororbia~\footnotemark[2]\\
    \texttt{ago@cs.rit.edu} \\
    \And
    Travis Desell~\footnotemark[2]\\
    \texttt{tjdvse@rit.edu} \\
}

\renewcommand{\shorttitle}{BP-free CANTS}

\hypersetup{
pdftitle={A template for the arxiv style},
pdfsubject={q-bio.NC, q-bio.QM},
pdfauthor={David S.~Hippocampus, Elias D.~Striatum},
pdfkeywords={First keyword, Second keyword, More},
}

\begin{document}
\maketitle
\begin{abstract}
Continuous Ant-based Topology Search (CANTS) is a previously introduced novel nature-inspired neural architecture search (NAS) algorithm that is based on ant colony optimization (ACO). CANTS utilizes a continuous search space to indirectly-encode a neural architecture search space. Synthetic ant agents explore CANTS' continuous search space based on the density and distribution of pheromones, strongly inspired by how ants move in the real world. 
This continuous search space allows CANTS to automate the design of artificial neural networks (ANNs) of any size, removing a key limitation inherent to many current NAS algorithms that must operate within structures of a size that is predetermined by the user. This work expands CANTS by adding a fourth dimension to its search space representing potential neural synaptic weights. Adding this extra dimension allows CANTS agents to optimize both the architecture as well as the weights of an ANN without applying backpropagation (BP), which leads to a significant reduction in the time consumed in the optimization process: {
% \color{red}
at least an average of $96\%$ less time consumption with very competitive optimization performance, if not better}. The experiments of this study - using real-world data - demonstrate that the \textbf{\textit{\mbox{BP-Free~CANTS}}} algorithm exhibits highly competitive performance compared to both CANTS and ANTS while requiring significantly less operation time.

% This work introduces a novel, nature-inspired neural architecture search (NAS) algorithm based on ant colony optimization, Continuous Ant-based Neural Topology Search (CANTS), which utilizes synthetic ants that move over a continuous search space based on the density and distribution of pheromones, strongly inspired by how ants move in the real world. The paths taken by the ant agents through the search space are utilized to construct artificial neural networks (ANNs). This continuous search space allows CANTS to automate the design of ANNs of any size, removing a key limitation inherent to many current NAS algorithms that must operate within structures of a size predetermined by the user. CANTS employs a distributed asynchronous strategy which allows it to scale to large-scale high performance computing resources, works with a variety of recurrent memory cell structures, and uses of a communal weight sharing strategy to reduce training time. The proposed procedure is evaluated on three real-world, time series prediction problems in the field of power systems and compared to two state-of-the-art algorithms. 
% Results show that CANTS is able to provide improved or competitive results on all of these problems while also being easier to use, requiring half the number of user-specified hyper-parameters.
%\keywords{Ant Colony Optimization \and Artificial Neural Network \and Neural Architecture Search}

\end{abstract}

% \begin{highlights}
% \item A robust metaheuristic algorithm that generates an indirect encoding of the neural search space to a continuous 4D search space.
% \item Backpropagation-free swarm intelligence NE offers a highly time efficient NeuroEvolution  method that effectively compete with backpropagation-NAS.
% \item Vast and unrestricted continuous search space helps in avoiding neural structural optimization local-minima-traps.
% \item CANTS is nature inspired and can be scaled up.
% \item The asynchronous design of CANTS significantly accelerates the optimization process. 
% \end{highlights}

% \begin{keyword}
%   NeuroEvolution \sep Neural Architecture Search \sep Swarm Intelligence \sep Genetic Evolution \sep Ant Colony Optimization \sep Time-series Prediction

% \end{keyword}

\section{Introduction}
\label{sec:introduction}

Hand-crafting artificial neural network architectures has been an obstacle in the advancement of machine learning (ML) as it is time-consuming, prone to trial-and-error, and requires significant domain expertise from model architects~\cite{zoph2016neural}. Exacerbating the problem, even slight alterations to problem-specific meta-parameters or topological features can lead to degradation in a model's performance~\cite{erkaymaz2014impact,Barna1990}.
% However, changing even a few problem-specific meta-parameters can lead to poor generalization upon committing to a specific topology~\cite{erkaymaz2014impact,Barna1990}.
This leads to the need for problem-specific model-optimization, however, the optimization of deep neural networks (DNN), with potentially millions of structural elements and a large number of hyperparameters, is considered an NP hard problem and limited by computational resources~\cite{a13030067}. Optimal architectures cannot be directly obtained by applying a continuous function because unlike the optimization of neural network parameters (weights) using a loss-function, an explicit function to measure the architecture-optimization is not available~\cite{liu2021survey}, due to the combinatorial and non-differentiable nature of architecture possibilities. Therefore, a number of meta-heuristic based neural architecture search (NAS)~\cite{elsken2018neural,liu2018darts,pham2018efficient,xie2018snas,luo2018neural,zoph2016neural} and neuroevolution (NE)~\cite{stanley2019designing,darwish2020survey} methods have been developed to automate the process of ANN design. Recently, nature-inspired neural architecture search (NI-NAS) algorithms have shown increasing promise, including the Artificial Bee Colony (ABC)~\cite{horng2017fine},  Bat~\cite{yang2010new}, Firefly~\cite{yang2010nature}, and Cuckoo Search~\cite{leke2017deep} algorithms.

Ant colony optimization (ACO) - originally introduced as a graph optimization method~\cite{dorigo1996ant} - has proven itself to be a successful NI-NAS strategy. The reason ACO operates as a strongly feasible NAS method is rooted in the concept of its application as a graph optimization method, providing impressive results~\cite{manfrin2006parallel,bianchi2002ant,dorigo1997ant,dorigo1997ant_,dorigo1996ant,dorigo1992optimization,dorigo2006ant,peake2019scaling}. As the structure of neural networks themselves are in essence directed graphs, this makes ACO well-suited towards the problem of NAS.

Initially, ACO for NAS was limited to small Jordan and Elman RNN neural structures ~\cite{desell2015evolving} or was used to select the network inputs~\cite{mavrovouniotis2013evolving}. ACO was also used for optimizing the synaptic connections (weights) within RNN memory cell structures~\cite{elsaid2018optimizing} and then later expanded to optimize entire RNN architectures within an algorithmic framework called Ant-based Neural Topology Search (ANTS)~\cite{elsaid2020ant}.
% Ant colony optimization (ACO)~\cite{dorigo1996ant} is a particularly successful NI-NAS strategy that has been shown to be quite powerful when automating the design of recurrent neural networks (RNNs). Originally, ACO for NAS was limited to small structures based on Jordan and Elman RNNs~\cite{desell2015evolving} or was used as a process for reducing the number of network inputs~\cite{mavrovouniotis2013evolving}. Later work proposed generalizations of ACO for optimizing the synaptic connections within RNN memory cell structures~\cite{elsaid2018optimizing} and even entire RNN architectures within an algorithmic framework called Ant-based Neural Topology Search (ANTS)~\cite{elsaid2020ant}. 
% In the ANTS process, ants traverse a single massively-connected ``superstructure'', searching for optimal RNN sub-networks which involves connectivity between RNN nodes both in terms of structure, \ie, all possible feed forward connections, and in time, \ie, all possible recurrent synapses that span many different time delays. This approach shares similarity to NAS methods in ANN cell and architecture design~\cite{liu2018darts,pham2018efficient,xie2018snas,cai2018proxylessnas,
%single/few shot methods%
% guo2020single,bender2018understanding,dong2019one,zhao2020few}, which operate within a limited search space, generating cells or architectures with a pre-determined maximum number of nodes and edges~\cite{elsken2018neural}.

ANTS utilizes a massively connected neural structure as a discrete structural search space. Although the approach has shown success, the main critique for the strategy is the limitation of discreteness of the search space. Other NAS methods are mostly evolutionary-based; instead of operating within fixed bounds, they are constructive and continually add and removing nodes/edges during the evolutionary process~\cite{stanley2002evolving,miikkulainen2019evolving,ororbia2019examm}.
Although that constructive NAS methods do not suffer from the limitations of a discreet search space -like ANTS - they are still more prone to falling into (early/poor) local minima or consume considerable time to evolve structures. 

% Alternatively, having to pre-specify bounds for the space of possible NAS-selected architectures often requires domain expertise and can lead to poorly performing or suboptimal networks if the bounds are incorrect, often  requiring many runs of varying bound values. 
This work introduces a novel ant colony inspired algorithm, the \linebreak \emph{BackPropagation-Free Continuous Ant-based Neural Topology Search} (BP-Free CANTS), which utilizes a continuous search domain that flexibly facilitates the design of ANNs of any size to address the aforementioned challenges. Synthetic continuous ant (\emph{cant}) agents roam explore the search space exploiting on the density and distribution of pheromone signals, emulating how ants swarm in the real world. The paths resulting from the agents exploration are used to construct nodes and edges of the RNN architectures. BP-Free CANTS (as well as the original CANTS algorithm~\cite{elsaid2021continuous}, which we will call BP-CANTS in this work for clarity) is a distributed, asynchronous algorithm, which facilitates scalable usage of high performance computing (HPC) resources, and also utilizes communal intelligence to reduce the amount of training required for candidate evolved networks. 
% Notably, the procedure allows for the selection of recurrent nodes from a suite of simple neurons and complex memory cells used in modern RNNs: $\Delta$-RNN units~\cite{ororbia2017diff}, GRUs~\cite{chung2014empirical}, LSTMs~\cite{hochreiter1997long}, MGUs~\cite{zhou2016minimal}, and UGRNNs~\cite{collins2016capacity}. 
% $\Delta$-RNN units~\cite{ororbia2017diff}, gated recurrent units (GRUs)~\cite{chung2014empirical}, long short-term memory cells (LSTMs)~\cite{hochreiter1997long}, minimal gated units (MGUs)~\cite{zhou2016minimal}, and update-gate RNN cells (UGRNNs)~\cite{collins2016capacity}. 

This work compares BP-Free CANTS to state-of-the-art benchmark NAS designing strategies for applied on RNN time series data prediction: ANTS~\cite{elsaid2020ant} and BP-CANTS~\cite{elsaid2021continuous}.  In addition to relaxing the requirement for \linebreak pre-determined architecture bounds, BP-Free CANTS is shown to yield results that improve upon ANTS and BP-CANTS at a significantly lower computational cost.  
%or are competitive to ANTS and EXAMM 
%while reducing the number of user specified hyperparameters from $16$ in both EXAMM and ANTS down to $8$ in CANTS. 
The BP-Free CANTS algorithm also provides an advancement to the field of NI-NAS is that it is able to efficiently both select for high performing architectures while at the same time finding well performing weights \emph{without backpropagation}. While ACO has been applied to continuous domain problems before~\cite{socha2008ant,kuhn2002ant,xiao2011hybrid,gupta2014transistor,bilchev1995ant}, to the authors' knowledge, BP-FREE CANTS is one of the first to simulate and apply the movements of ants through a unbounded 4D continuous space to search for optimal neural parameters and architectures. 
BP-Free CANTS has the capacity to optimize the neural topology and the neural synaptic parameters without applying the time consuming backpropagation and gradient-descent process required for memetic algorithms such as CANTS or other NE-based methods, further reducing required computational power and time consumption. BP-FREE CANTS also provides means for its cants agents to evolve during the optimization process, which enhances the chances of reaching better performing models, while requiring fewer hyperparameters that are associated with cants' behavior.

\begin{algorithm}
	\footnotesize
    \caption{Continuous Ant-guided Neural Topology Search Algorithm}\label{alg:cants_pseudo}
    \begin{algorithmic}
		\Procedure{$Work Assignor$}{}
			\LineComment{Build 4D space with:}
            \LineComment{inputs at $y_{axis}=0$}
            \LineComment{output at $y_{axis}=1$}
			\LineComment{$z_{axis}$: recurrent time steps}
            \LineComment{$x_{axis}$: x component of neuron position}
			\State $space \gets \textbf{new}~SearchSpace$
			\For {$i \gets 1 \dots optimization\_rounds$}
				\State $nn \gets Swarm( )$
				\State $to\_assignee (nn_{new}, assignee.id)$
				\State $nn, fitness \gets fitness\_from\_assignee( )$
				\If { $nn\_{fit}< population.worst\_member$ }
					\State $population.pop ( population.worst\_member)$
					\State $population.join ( nn )$
					\State $DopositePheromone ( nn )$
				\EndIf
                \For {$cant \gets 1 \dots no\_cants$}
				    \State $cant.evovle(fitness)$
			    \EndFor
			\EndFor
		\EndProcedure
		
		\Procedure{$Assignee$}{}
			\State  $nn \gets from\_assignor ()$		 
			\State $fit \gets train\_evaluate\_nn ( nn )$
			\State $to\_assignor ( nn, fit )$
		\EndProcedure

		\Procedure {$Swarm$} {}
			\For {$cant \gets 1 \dots no\_cants$}
				\State $TakePath( cant )$
			\EndFor
			\LineComment { Cluster paths vertices using DBscan  }
			\State { $segments \gets PathsDBS(cants)$ }
			\LineComment {Construct architecture from segments}
			\State $rnn \gets BuildRNN(segments)$
			\Return $rnn$
		\EndProcedure

		\Procedure {$TakePath$} {$cant$}
			\LineComment {Input chosen discretely}
			\State $PickInput( cant )$
			\While { $cant.y_{curr} < 0.99$}
				\State $r \gets rand_{uni}(0, total\_pheromone - 1)$
				\State $cant.level_{curr}  \gets cant.go\_up$
				\If { $r > ant.explore\_intuition $ \textbf{or} $ space[cant.level_{curr}] $ not $ Empty  $ } 
					\State $spot \gets PutFootPrint(cant.radius_{sense})$
					\State $cant.path.add( spot )$
					\State $space.add( spot )$
				\Else
					\State $spot \gets locateCenterOfMass(cant.pos_{curr}, cant.radius_{sense})$
					\If { $spot$ not in $space[cant.level_{curr}]$}
						\State $cant\_path.add( spot )$
					\EndIf
				\EndIf
			\EndWhile
            \State $PickOutput(cant)$
		\EndProcedure

		        \algstore{myalg}
		    \end{algorithmic}
		\end{algorithm}

		\begin{algorithm}
		 			\footnotesize{}
		 		    \begin{algorithmic}
		 		        \algrestore{myalg}
		
		\Procedure {$cant.Evolve$} {$fitness$}
            \State $B = cant.behavior_{curr}$
            \If {$fitness < behaviors.worst$}
                \State $B_{new} \gets  (cant.rate_{explore\_exploit}, cant.radius_{sense}, cant.r1, cant.r2)$
                \State $behaviors.join(B_{new})$
            \EndIf
            % \State $r \gets r ← random_{uni}(0, 1)$
            \If {$cant.best\_bahaviors<10$ \textbf{or} $ random_{uni}(0, 1)<\sigma_{mutation}$}
                \State $B \gets Mutate(B)$
            \Else
                \State $B \gets CrossOver(B, cant.behavior_{best\_1}, cant.behavior_{best\_2})$
            \EndIf
        \EndProcedure

        \Procedure{$Mutate$}{$behavior$}
            \State $behavior.rate_{explore\_exploit} \gets random(0,1)$
            \State $behavior.rate_{sense} \gets random(0,1)$;
            \State $behavior.r1 \gets random_{uni}(-1, 1)$
            \State $behavior.r2 \gets random_{uni}(-1, 1)$
        \EndProcedure

        \Procedure{$CrossOver$} {$behavior, behavior1, behavior2$}
            \State $behavior \gets \big((behavior2 - behavoir1) \times random(0,1) \big) + behavior1$ 
        \EndProcedure
        
		\Procedure {$PickInput$} {$cant$} 
		    \LineComment{Probabilistically with pheromones density}
			\State $total\_pheromone \gets \textbf{sum}(inputs.pheromones)$
			\State $r \gets random_{uni}(0, total\_pheromone - 1)$
			\State $cant.input \gets 0$
			\While {$r > 0$}:
	            \If {$r < inputs.pheromones[cant.input]$}
	                \State $cant.input \gets + 1$
	                \State \textbf{break}
	            \Else
	                \State $r \gets r - inputs.pheromones[cant.input]$
	                \State $cant.input \pluseq 1$
	            \EndIf
			\EndWhile
		\EndProcedure

		\Procedure {$PickOutput$} {$cant$} 
		    \LineComment{Probabilistically with pheromones density}
			\State $total\_pheromone \gets \textbf{sum}(outputs.pheromones)$
			\State $r \gets random_{uni}(0, pheromone\_sum - 1)$
			\State $ant.input \gets 0$
			\While {$r > 0$}:
	            \If {$r < outputs.pheromones[cant.output]$}
	                \State $cant.output \gets 1$
	                \State \textbf{break}
	            \Else
	                \State $r \gets r - outputs.pheromones[cant.output]$
	                \State $cant.output \pluseq 1$
	            \EndIf
			\EndWhile
		\EndProcedure

		%         \algstore{myalg}
		%     \end{algorithmic}
		% \end{algorithm}

		% \begin{algorithm}
		%  			\footnotesize{}
		%  		    \begin{algorithmic}
		%  		        \algrestore{myalg}

		\Procedure {$PathsDBS$} {$cants$} 
			\For {$cant \gets 1 \dots cants\_count$}
				\For {$spot \gets 1 \dots cant\_path$}
					\State {$segments[cant].insert(PickSpot(spot))$}
				\EndFor
			\EndFor
			\Return $segments$
		\EndProcedure

		\Procedure { $PickSpot$} {$spot$}
			\State $ [node, cluster_{spots}] \gets PathDBS(spot, space[point.level])$
			\State $node.edges_{fan_out}.weight \gets (AverageWeights(cluster_{spots}))$
			\State $space.add(node)$
			\Return $node$
		\EndProcedure

		\Procedure {$DepositPheromone$} {$nn$} 
			\ForEach {$node \in nn.nodes$} 
				\State {$space[ node ].pheromone \pluseq const $}
				\State {$space[ node ].weight \gets Average(node.weight, space[ node ].weight)$}
				\If { $space[ node ].pheromone > THRESHOLD$ }
					\State { $space[ node ].pheromone = PHEROMONE$ }
				\EndIf 
			\EndFor
		\EndProcedure
	
\end{algorithmic}
\end{algorithm}	

\section{Methodology}
\label{sec:method}

\begin{figure}[!t]
    \includegraphics[clip,width=0.75\textwidth, height=.4\textwidth]{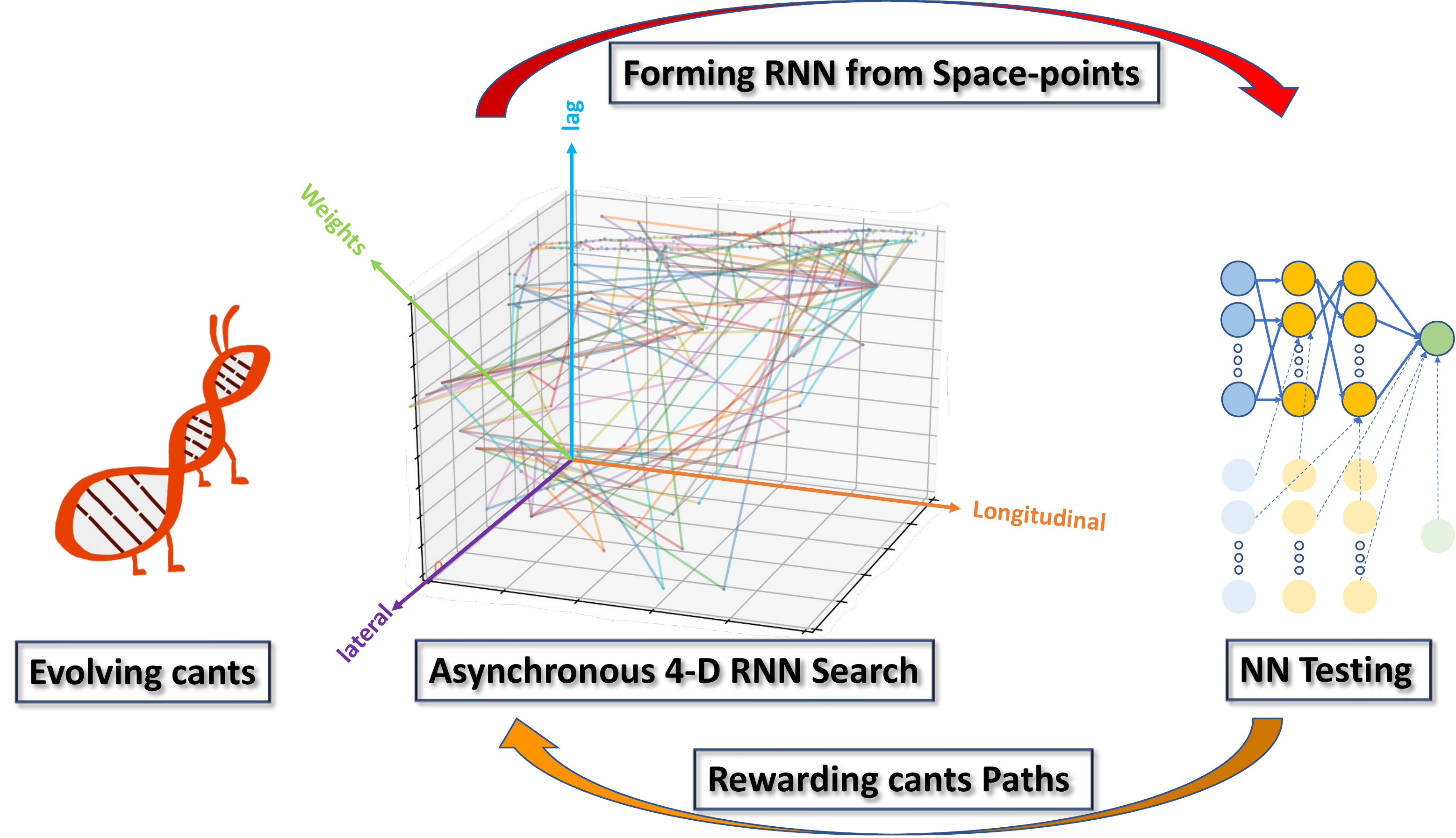}
    \caption{\centering \textit{The BP-free CANTS schematic graph.}\label{fig:cants_sketch}}
    \vspace{-0.2cm}
\end{figure}

\begin{wrapfigure}[]{r}{0.6\textwidth}
% \begin{figure}
    \vspace{-10pt}
    \centering
    \includegraphics[width=.98\textwidth, height=.5\textwidth]{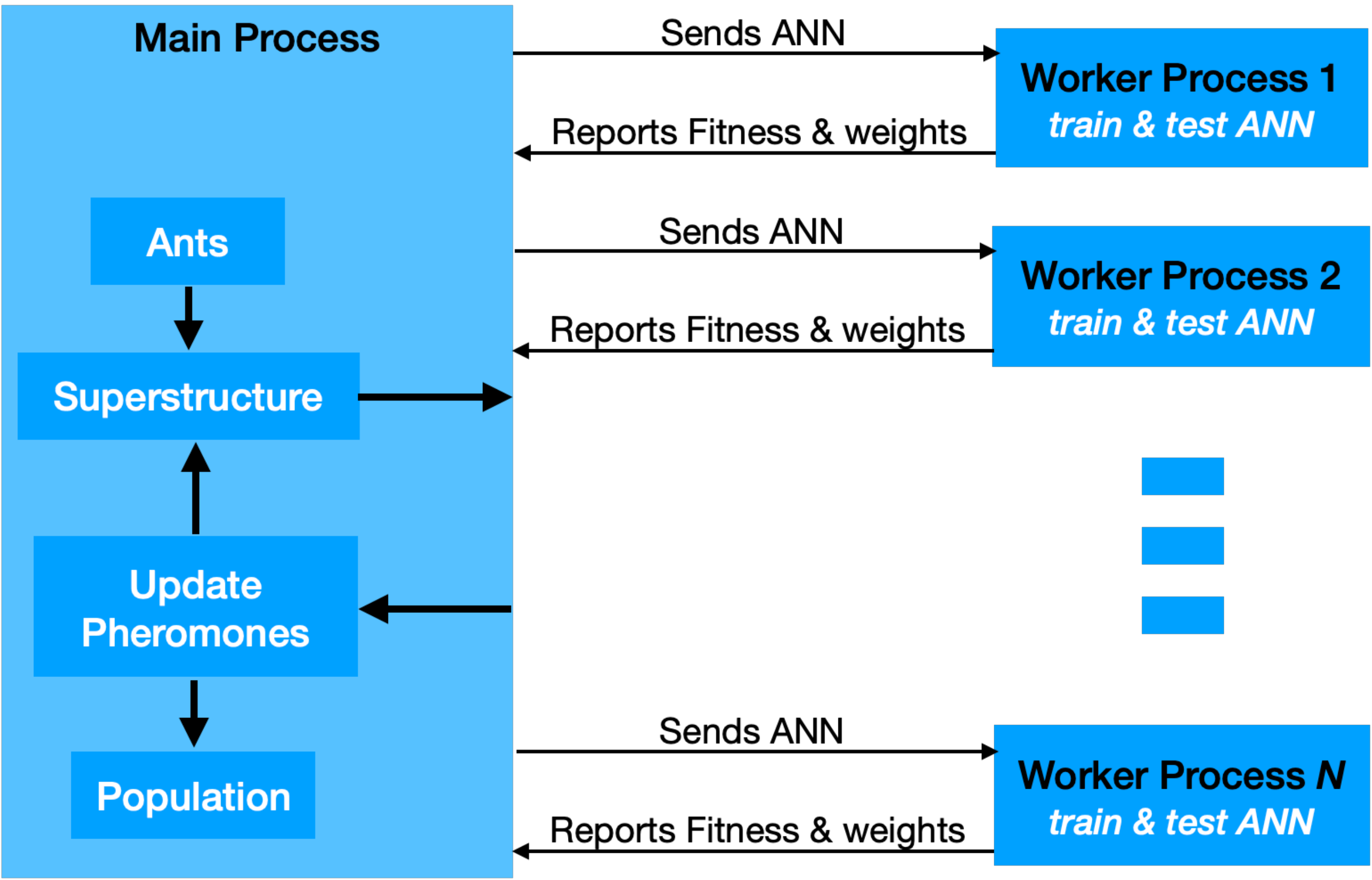}
    \caption{\textit{The CANTS asynchronous design.}}
    \label{fig:async_ants}
% \end{figure}
\vspace{-0.2cm}
\end{wrapfigure}  

Figure~\ref{fig:cants_sketch} presents a abstract-level illustration for how the BP-Free CANTS algorithm works.
An asynchronous, distributed ``work-stealing'' strategy is applied (see pseudo-code in Algorithm~\ref{alg:cants_pseudo}) for scalable use on HPC clusters\footnote{Source code is implemented in Python and is offered as an open-source project on \href{https://github.com/a-elsaid/CANTS_public.git}{https://github.com/a-elsaid/CANTS\_public.git}}. When receiving a request from a worker process (assignee), the manager (assignor) process generates new architectures. When requesting a new architecture, worker reports the fitness of the architecture it trained and evaluated. 
A fixed population of the best RNNs discovered by the workers is maintained by the manager process. 
The manager process also rewards the paths that the ant agents took to generate the RNN architecture by depositing pheromones in the continuous search space it manages. This computation design allows workers to evaluate and report the performance of the a single generated architecture at a time, without blocking waiting on results from other workers, offering a naturally load-balanced algorithm.
% This strategy allows workers to complete the training of the generated RNNs at whatever speed they are capable of, yielding an algorithm that is naturally load-balanced. 
Additionally, this allows BP-Free CANTS increased scalability over synchronous parallel evolution strategies because it can support a number of worker processes greater than the population size.
% Unlike synchronous parallel evolutionary strategies, BP-Free CANTS scales up to any number of available processors, supporting population sizes that are independent of processor availability. 
The population managed by the manager process houses the best fitness (mean squared error on validation data) of the discovered RNNs reported by workers. This population is held at a fixed size and updated as in steady-state evolutionary algorithms, where the worst member of the population is removed and replaced any newly discovered RNN that performs better.
% When the resulting fitness (mean squared error over validation data) of candidate RNNs is reported to the work generator process, if the candidate RNN is better than the worst RNN in the population, then the worst RNN is removed and the candidate is added. Note that the saved pheromone placement points for the candidate are incremented in the continuous search space. 

%\zimeng{This sentence needs revision: Candidate RNNs are generated using a 3-dimensional search space is divided to planes or levels each representing a time-step, with the current time-step at the top layer and lower layers representing preceding time steps}. 

%reworded to be more concise
%-- this constraint is imposed because the movement of ants between the layers in our algorithm's search space represents the forward propagation across time-steps, hence it is only possible to propagate information from a previous time step ($t-k$) up to and including the current step $t$ but not the reverse, since this would imply carrying unknown, future signals backwards. 

Candidate neural architectures are sampled from a search space comprised of stacked 2D continuous planes, where each 2D plane represents a particular time step $t$ (see Figure \ref{fig:cants_move_explore}), and crucially in BP-Free CANTS,  the fourth dimension of a given point in the search space represents the weight value of that point.  
The input features at each time step, represented as the input nodes, are uniformly distributed at the zero point of one of the axes of the search space. 
A synthetic continuous ant agent (or \emph{cant}) discretely selects an input node position to start its path to an output node. The cant then moves on its path through the continuous space based on the current density and distribution of pheromone traces. 
On a given plane (lag-level) in the search space, a cant can only move forward toward the side of the outputs, or jump up to any lag-level above, but it is not permitted to moving to lower lag-levels in the stack, or to move backwards on the same lag-level. As paths (recurrent connections) moving up the stack represent passing information from neurons at previous time step to neurons at a future time step, the reverse would dictates passing unavailable future information to a previous time step of the RNN, which is not possible. Although cants are restricted to move only forward on a given plane, they are allowed to move backwards when they do a jump to a higher lag-level since recurrent connections can feed information from an architecture layer-level to lower layer-level that has a higher lag-level. 
% Cants can only move forward on the stack level they are currently on and can move up to any plane above it. They are restricted from moving down the stack -- while connections moving up the stack represent passing information from a previous time step to a future time step, the reverse would require passing unknown future data to a previous time step of the RNN which is not permitted. While ants only move forward on a given plane, they are permitted to move backward when moving to a plane higher on the stack since many RNNs have recurrent connections that feed into earlier nodes in the network. 
This enforced (overall) forward movement on the planes (through layer-level) and upward on the stack (through lag-levels) ensures the continuous progress of cants towards the outputs, and alleviates the cycles in the search space. 

Figure~\ref{fig:cants_move} shows scenarios of cants movement through the search space starting from an input to an output. The shown movements illustrate cants exploration of new spots in the search space, cants exploitation of previously searched areas, attracted by previous pheromone deposits, and how cant-paths are translated into a final candidate architecture (refer to~\cite{elsaid2021continuous} for more details).

%The software developed that implements our CANTS procedure also provides a replay visualization tool so that traces from a run can be visualized to see how pheromones are deposited and how RNNs are generated, as shown in Figure~\ref{fig:cants_replay_visualization}.

%\begin{figure}
%    \centering
%    \includegraphics[width=1\textwidth, height=.95 \textheight]{./img/cants_activity_dia.pdf}
%    \caption{{\textit{High-level CANTS Work Flow}}}
%    \label{fig:cants_flowchart}
%\end{figure}

\begin{figure}
    \centering 
    \begin{tabular}{cc}
        \subfloat[][ \emph{cant picking $1^{st}$ spot (explore move)} ]{\includegraphics[width=0.5\textwidth, height=0.15\textheight]{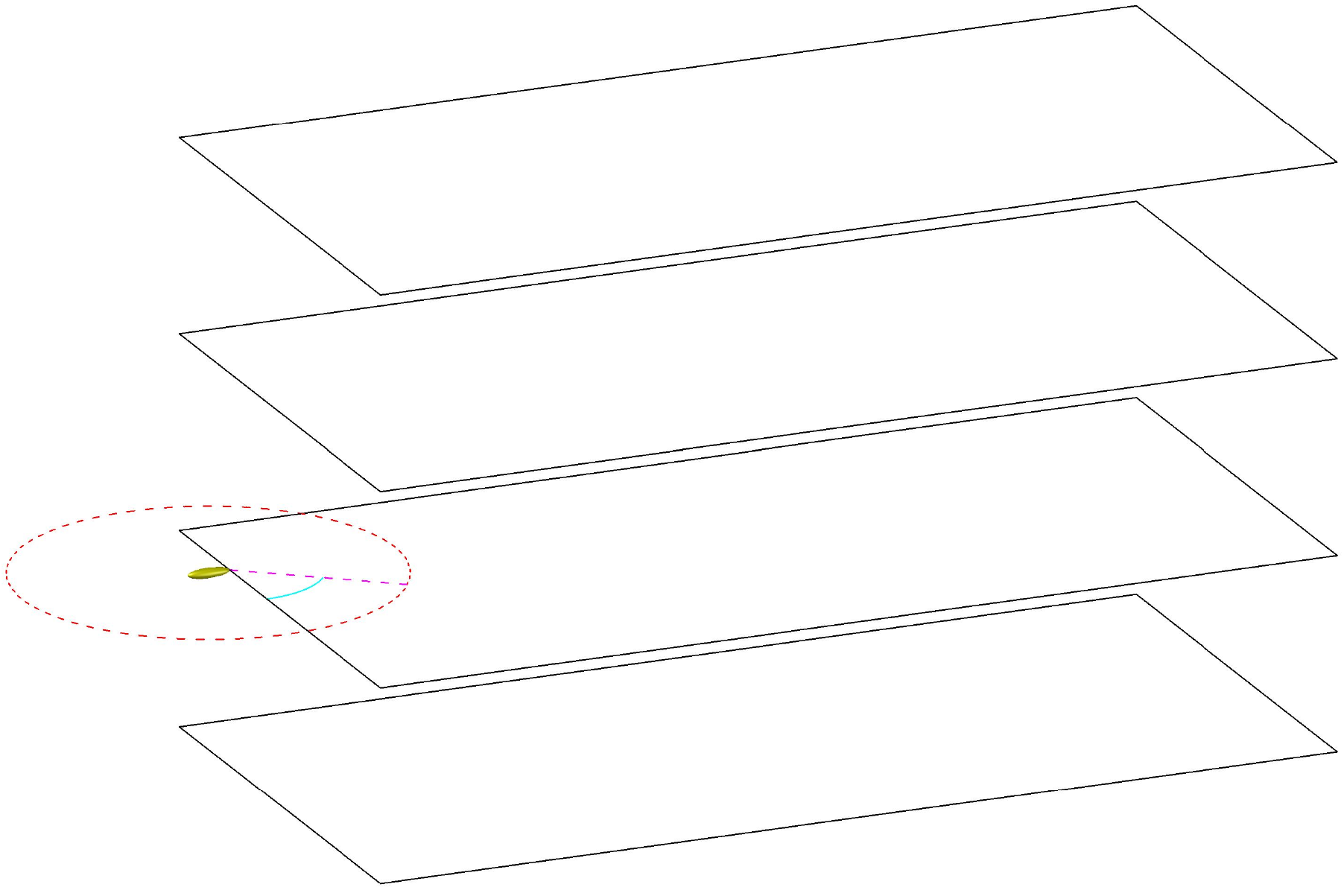} \label{fig:cants_move_explore}} &
        \subfloat[][ \emph{cant picking a spot (exploit move)}]{\includegraphics[width=0.475\textwidth, height=0.15\textheight]{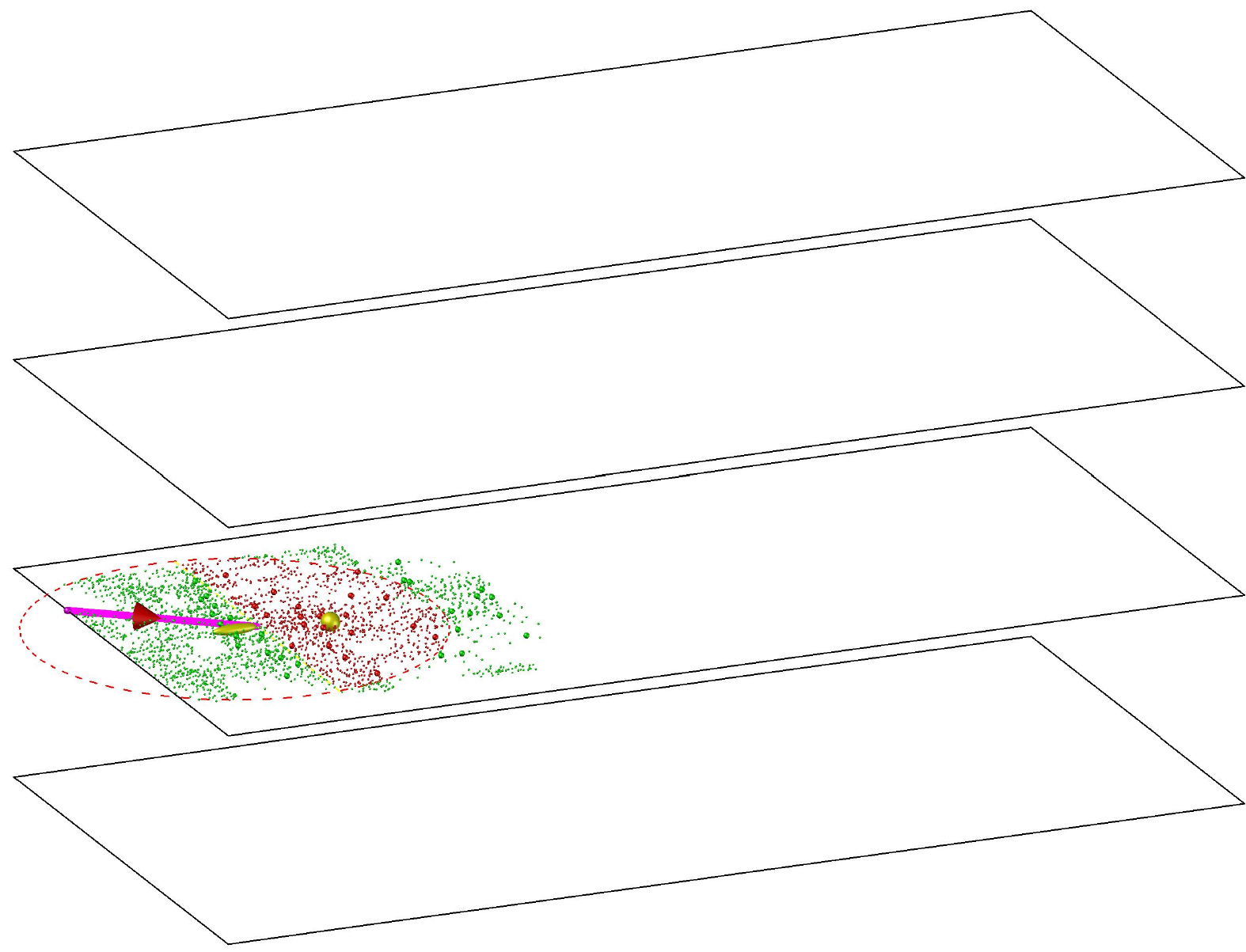} \label{fig:cants_move_exploit_1}} \\

        \subfloat[][ \emph{cant picking a spot in higher level}]{\includegraphics[width=0.475\textwidth, height=0.15\textheight]{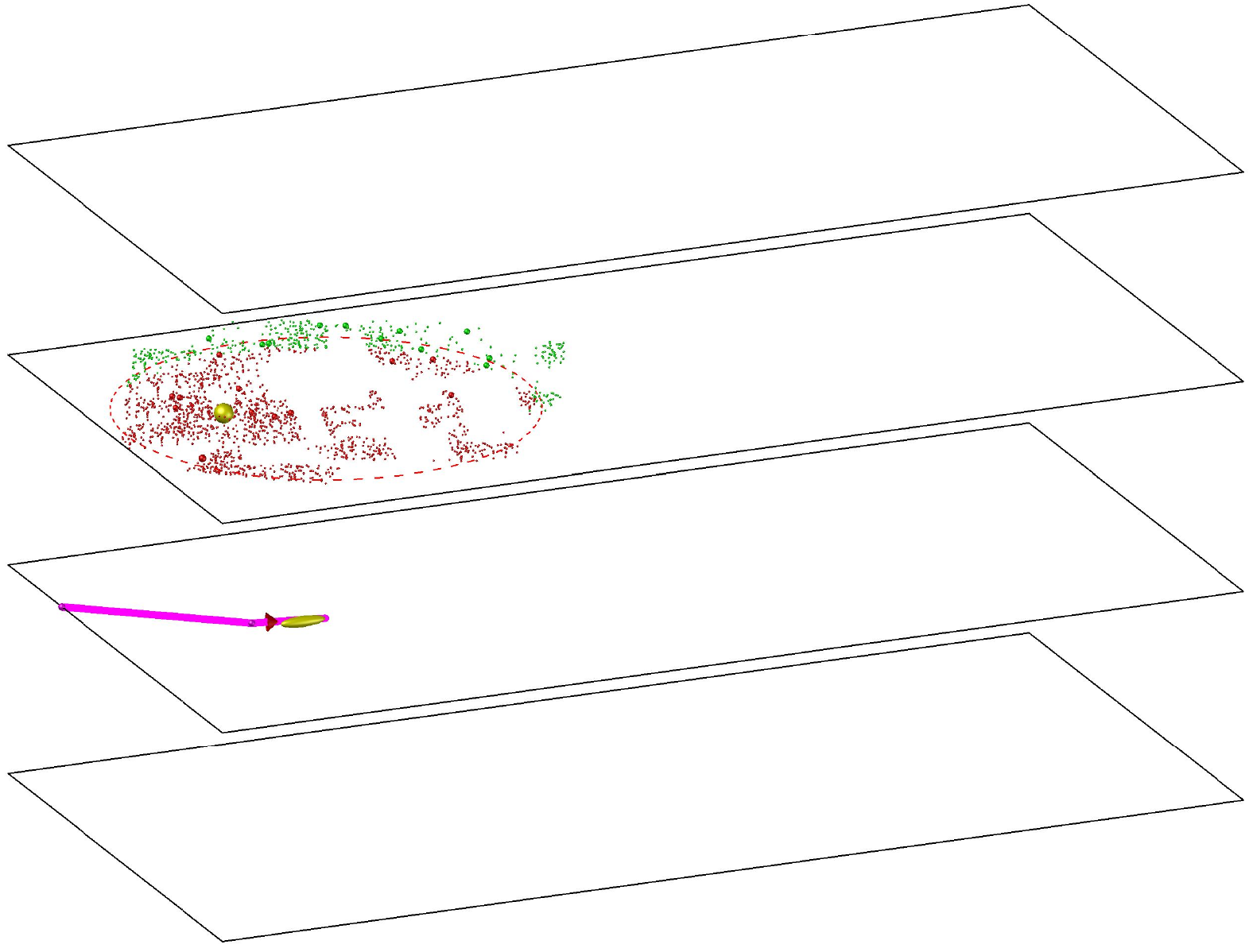} \label{fig:cants_move_exploit_2}} &
        \subfloat[][\emph{moving to the new spot}]{\includegraphics[width=0.475\textwidth, height=0.15\textheight]{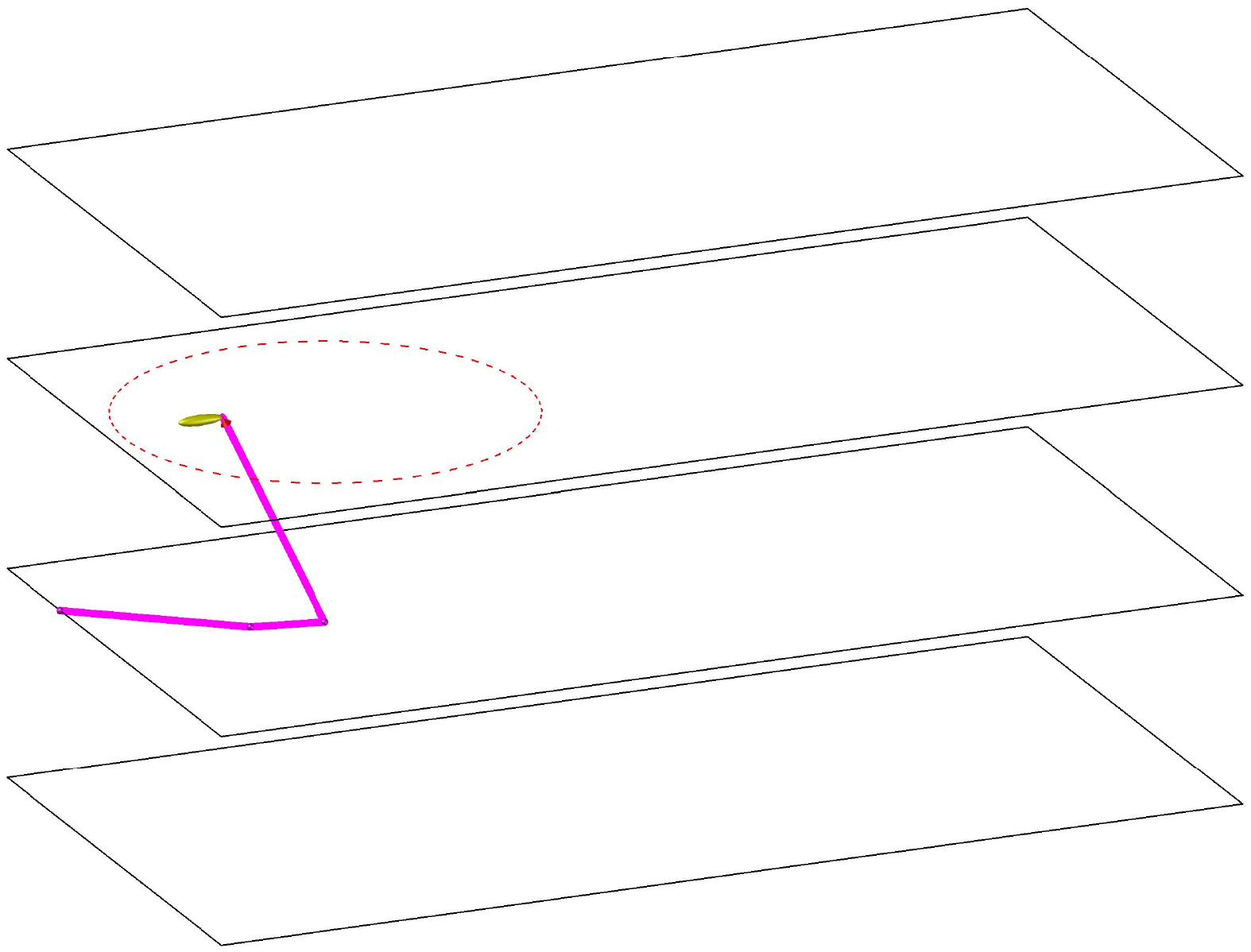} \label{fig:cants_move_05}} \\
        
        \subfloat[][ \emph{another spot}]{\includegraphics[width=0.475\textwidth, height=0.15\textheight]{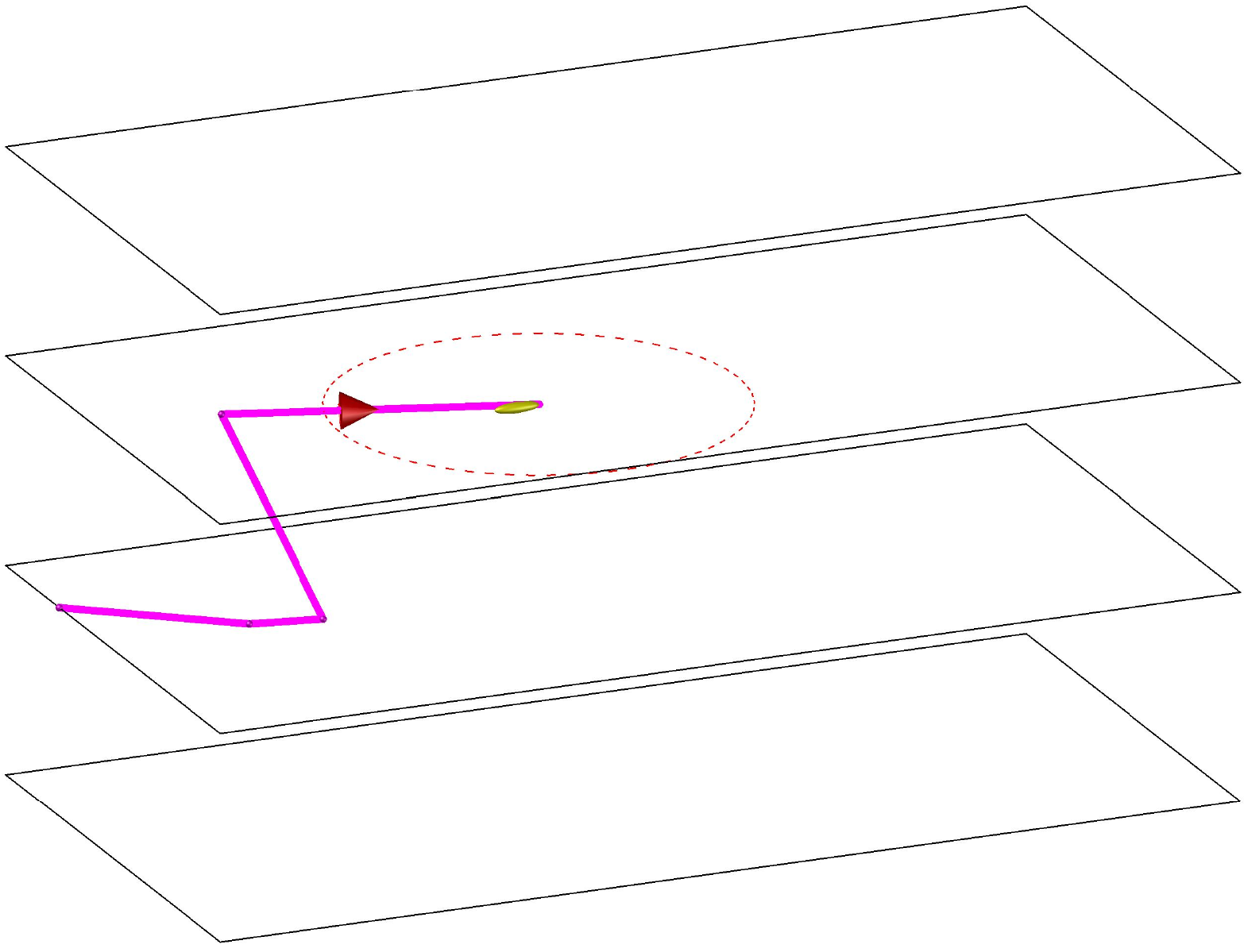} \label{fig:cants_move_06}} 
        &
        \subfloat[][ \emph{picking an output}]{\includegraphics[width=0.475\textwidth,height=0.15\textheight]{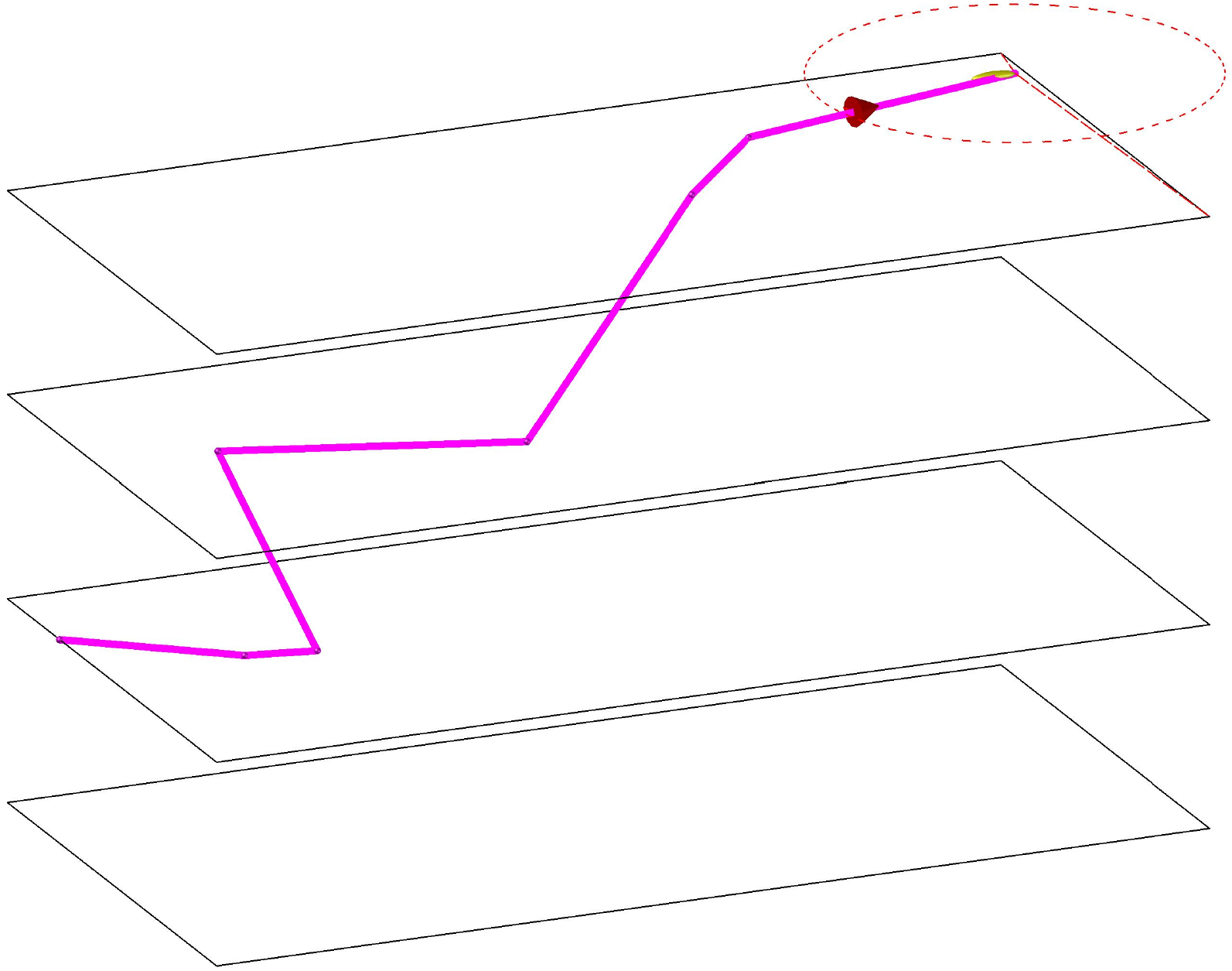} \label{fig:cants_move_pick_output}} 
        \\
        \subfloat[][\emph{cants paths}\label{fig:ants_multi_path}]{\includegraphics[width=0.475\textwidth,height=0.25\textheight]{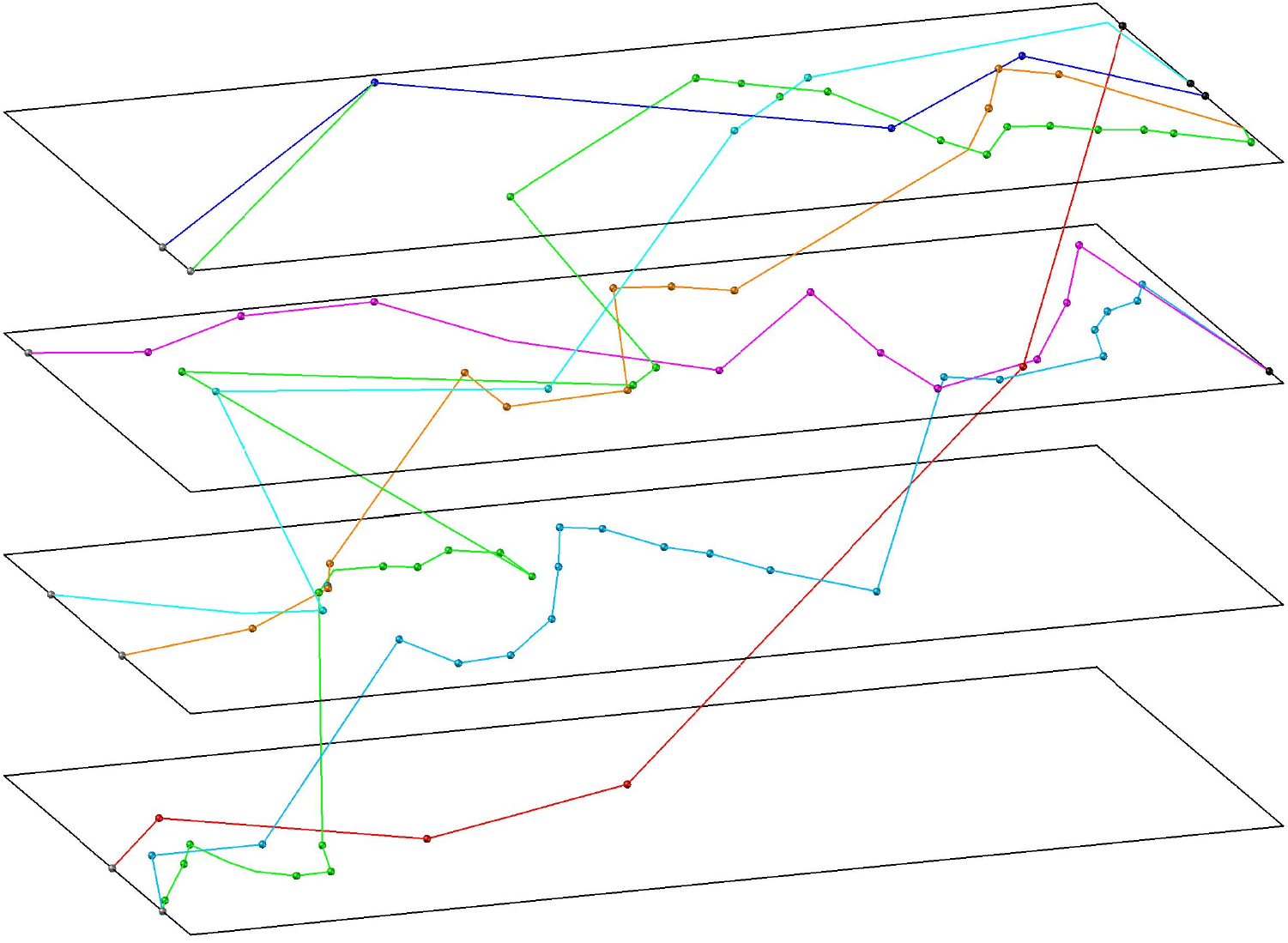} \label{fig:cants_move_10}} 
        &
        \subfloat[][\emph{cants path after condensation} \label{fig:ants_condensed_path}]{\includegraphics[width=0.45\textwidth, height=0.25\textheight]{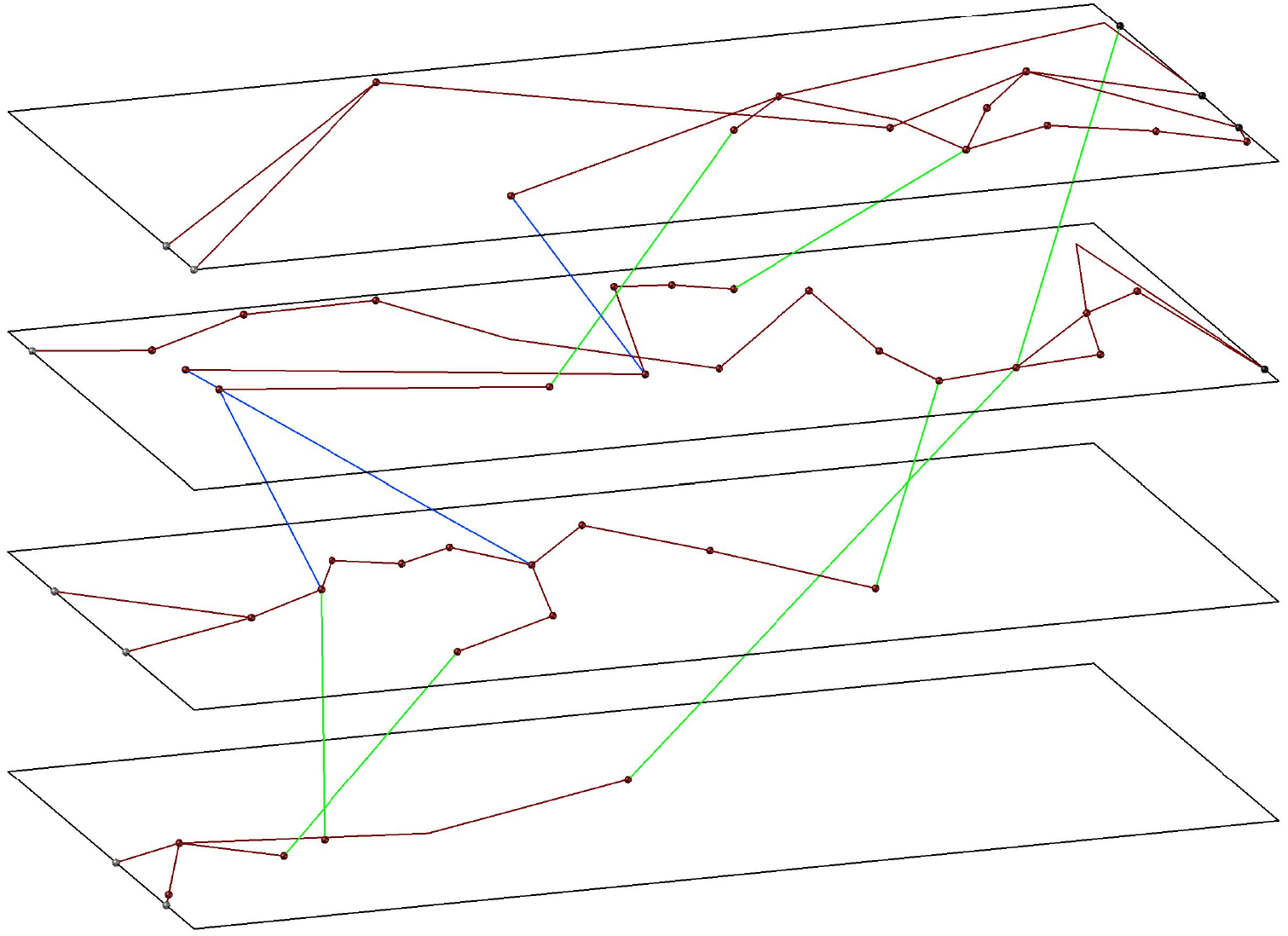} \label{fig:cants_move_11}} \\
    \end{tabular}
    \caption[]{\textit{\textbf{CANTS paths \& Architecture Building}}}
\end{figure}

\begin{figure}[h]
        \ContinuedFloat 
    \centering 
    \begin{minipage}{0.48\textwidth}
    % \subfloat[][]{\includegraphics[width=0.95\textwidth,height=0.2\textheight]{./img/cants_movements_8-Model.pdf} \label{fig:cants_move_08}} 
    % \\
    \subfloat[][\emph{2D projection of a cant path}]{\includegraphics[width=0.95\textwidth,height=0.39\textheight]{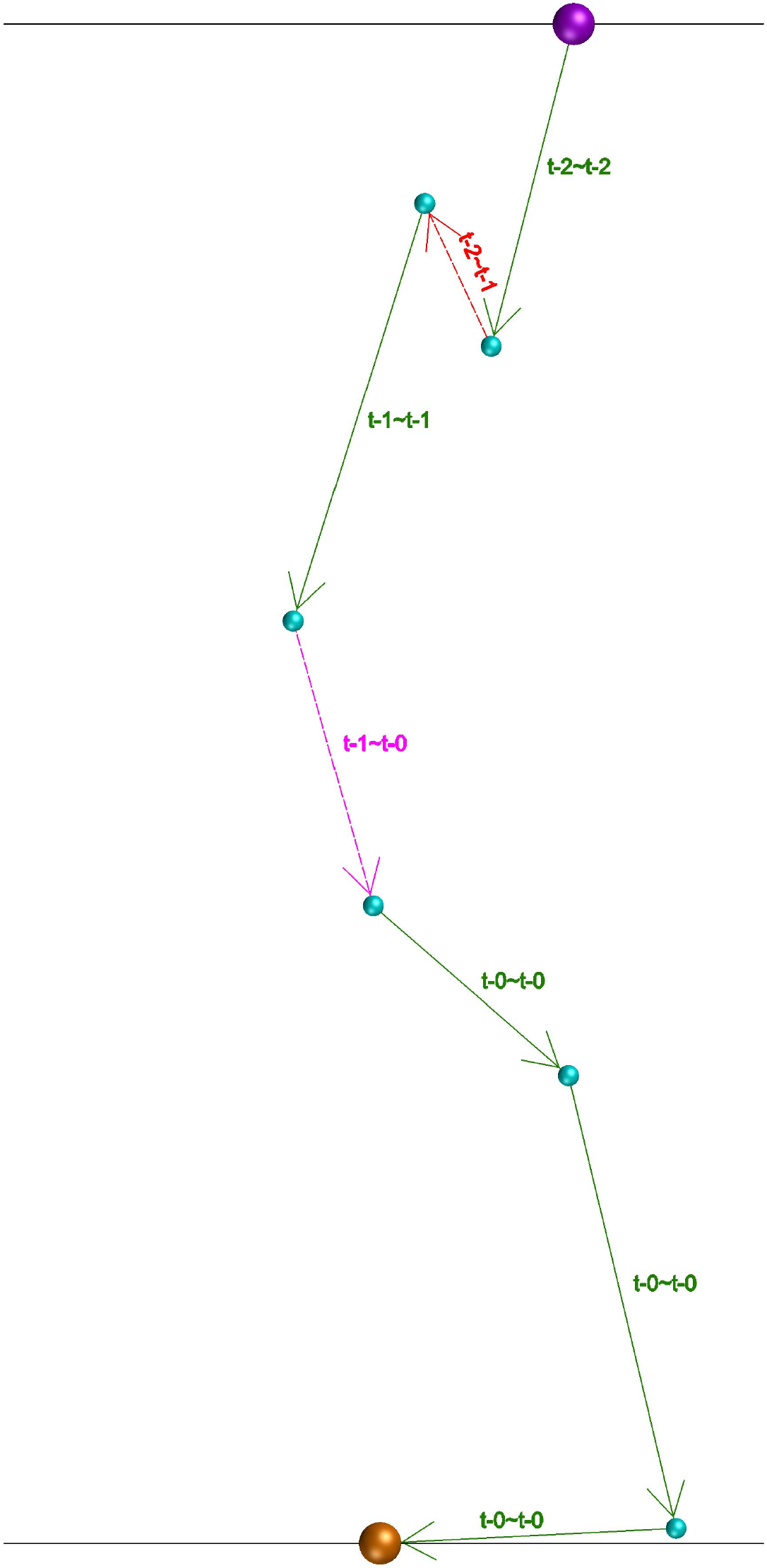} \label{fig:cants_move_09}}
    \end{minipage}%
    \begin{minipage}{0.48\textwidth}
            \subfloat[][\emph{2D projection of cants paths}]{\includegraphics[width=0.9\textwidth,height=0.39\textheight]{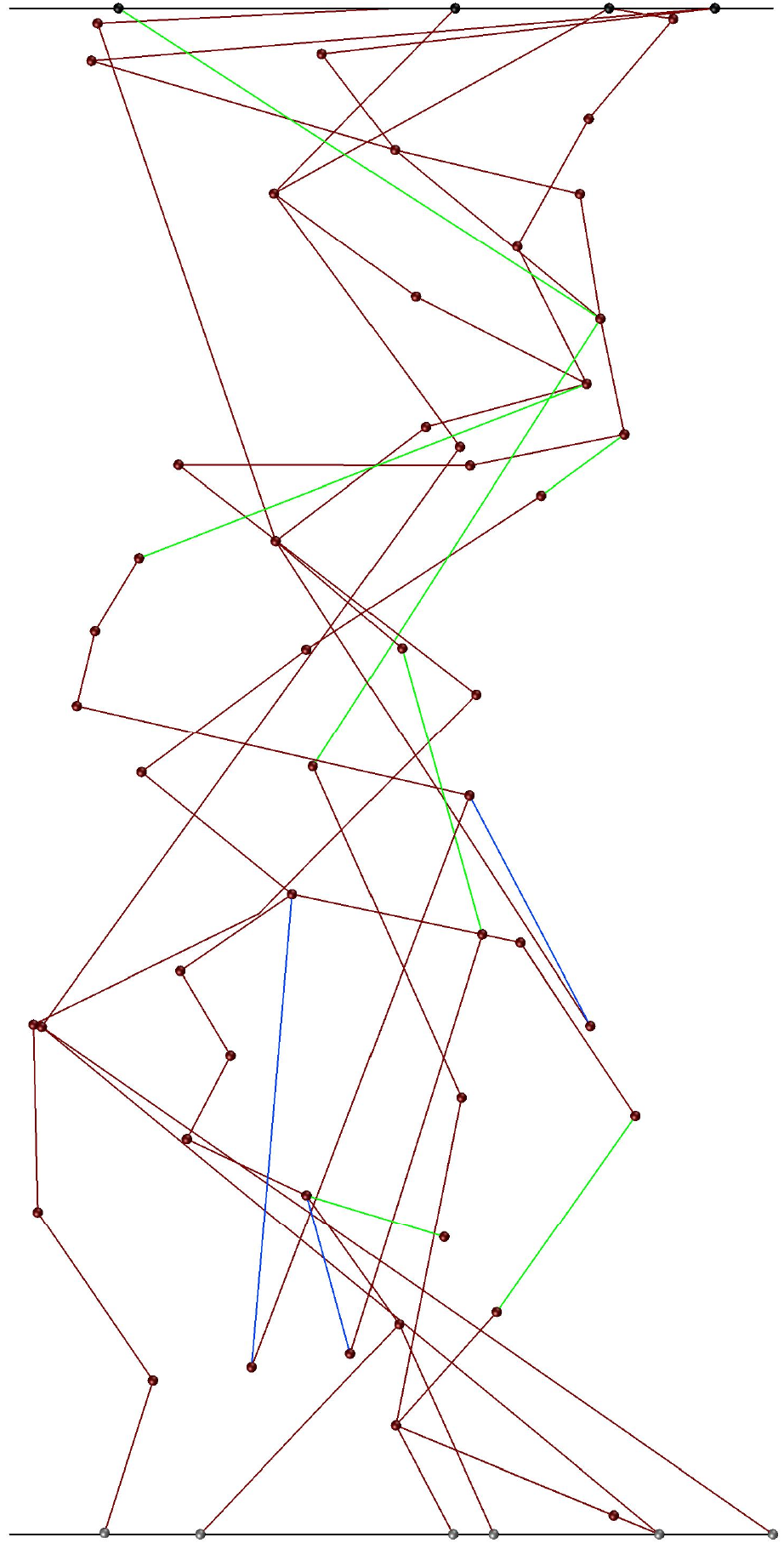} \label{fig:cants_move_12}}
    \end{minipage}
    \caption[]{\textit{\textbf{CANTS paths \& Architecture Building (continued):}\label{fig:cants_move}}
    % (i) An example cant's path is projected on one plane. 
    % (i) The cant picked its input point, starting at level $t_{-2}$, picked a node at $t_{-2}$ (green edge), picked a node at $t_{-1}$ (red backward recurrent edge), picked a node at $t_{-1}$ (green edge), picked a node at $t_{0}$ (magenta forward recurrent edge), picked a node at $t_0$ (green edge), picked a node at $t_0$ (green edge), and finally picked an output node at $t_{0}$ (green edge). (j) The final network is the final result of clustering the nodes and defining the connections between nodes in the same layer as red edges, and the connection between nodes and between layers as green forward recurrent edges or blue backward recurrent edges. The flow moves from the gray inputs at the bottom to the black outputs at the top.
    \label{fig:cants_move2}}
\end{figure}

% \caption[]{{\bf Cant path selection and network construction (continued):} (i) The cant's selections: start at level $t_{-2}$, input point, node at $t_{-2}$ ({\color{ForestGreen} edge}), node at $t_{-1}$ ({\color{red} backward recurrent edge}), node at $t_{-1}$ ({\color{ForestGreen} edge}), node at $t_{0}$ ({\color{magenta} forward recurrent edge}), node at $t_0$ ({\color{ForestGreen} edge}), node at $t_0$ ({\color{ForestGreen} edge}), and output node at $t_{0}$ ({\color{ForestGreen} edge}). (k) The final network is the final result of clustering the nodes and defining the connections between nodes in the same layer as {\color{red}edges}, and the connection between nodes and between layers as {\color{green}forward recurrent edges} or {\color{blue}backward recurrent edges}. The flow moves from the {\color{gray} inputs} at the bottom to the {\color{black} outputs} at the top. \label{fig:cants_move2}}

\begin{comment}
\begin{figure}
    \centering
    \begin{tabular}{cc}
    \subfloat[{Initially generated RNN}\label{fig:frame1}]{
        \centering
        \includegraphics[width=.49\textwidth]{./img/frame_0.pdf}
    } &
    \subfloat[{$123^{rd}$ generated RNN}\label{fig:frame2}]{
        \centering
        \includegraphics[width=.49\textwidth]{./img/frame_123.pdf}
    }
    \caption{{A replay visualization of CANTS showing the pheromones and paths taken by ants through the complex, continuous search space.}}
    \label{fig:cants_replay_visualization}
    \end{tabular}
\end{figure}
\end{comment}

\paragraph{Cant Agent Input Node and Layer Selection:} 
Each level in the search space has a level-selection pheromone value, $p_l$, where $l$ is the level. These are initialized to $p_l = 2 * l$ where the top lag-level for the current time step is $l = 1$, the next lag-level for the first time lag is $l = 2$ and so on. A cant selects its starting lag-level according to the probability of starting at lag-level $l$ under $P(l) = \frac{p_l}{\Sigma^L_{l=1} p_l}$, where $L$ is the total number of lag-levels. This scheme encourages cants to start at lower lag-levels of the stack at the beginning of the search. After selecting a lag-level, the cant selects its input node in a similar fashion, based on the pheromones for each input node location on that lag-level. When a candidate RNN is inserted into the population, the lag-level pheromones for each lag-level, utilized by that RNN, are incremented.

\paragraph{Cant Agent Movement:} 
The movement of cants in the search space is a balance between exploitation and exploration, which is reached by imitating real-world ants, who follows clues to communicate information about a target. 
% To balance exploration with exploitation, cants behave similarly to real-world ants by following communication clues in order to reach to targets. 
When a cant takes a step, it first decides if it will climb to higher lag-level in the planes stack, which is done similarly to how they select the initial lag-level with levels options limited to only the higher planes up the stack.
% When a cant moves, it first decides if it will climb up to a higher (stack) level. This is done in the same manner as selecting its initial layer, except that it only selects between its current level and higher ones. 
When the cant determines if the next move will be a climb or not, it will make a choice to whether the move will be an exploration or exploitation step. 
The movement type (exploration or exploitation) is decided by the cant based on its exploration/exploitation parameter, which is initially generated using a uniform random distribution, and then later evolves through the generation of neural architectures, as one the agent's behaviors (see Section {\it Cants Evolution} below).
% After deciding if it will climb or not, the agent will then decide if it will explore or exploit. Cants randomly choose to exploit at a percentage equal to an exploitation parameter, $\epsilon$. 

In exploitation moves, \ie, following pheromone traces/clues, a cant will first sense the pheromone traces around, that is lying within its sensing radius, $\rho$.
% When a cant decides to exploit and follow pheromone traces, \ie, clues, it will start sensing the pheromone points around it, given a sensing radius, $\rho$. 
If the move is decided to be on the same lag-level, the cant will ignore previously-deposited pheromone that is behind it, otherwise, the whole pheromone distribution inside the sensing radius will be considered by the cant to determine the next location in its path. 
% If the cant choose to stay on the same level, it will only consider deposited pheromone that are in front of it (\ie, closer to the output nodes), otherwise, it will consider all the pheromones that are inside its sensing radius on the level it is moving to. 
The center-of-mass of the pheromone distribution inside the sensing radius is calculated by the cant as the location of the next step in its path.
% The cant then calculates the center of mass of the pheromones within this region using the point in the space it will move to. 
Cants' path points are saved by the their candidate generated RNN for later pheromone deposit, for future communication with other cants, if the RNN made it to the population of best discovered RNNs. 
% This point is then saved by the candidate RNN (as a point to potentially increment pheromone values) if the RNN is later to be inserted into the RNN population. 
Applying the center-of-mass in cants selection to their next move makes the placement of pheromone at individual points a peripherally effective factor in cant-to-cant communication compared to the concentration of pheromone in a region in the space, which resembles real-ants foraging in nature. 
% Since cants consider the center of mass of the pheromone values, the individual points of pheromone values are not the effective factor in cant-to-cant communication. Rather, it is the very concentration of the pheromone in a region of the space that more closely aligns with how real ants move in nature. 

When an exploration move is decided by a cant, it will pick a random point lying within its sensing radius for its next location. 
% When a cant instead decides that it will explore, it instead selects a random point that lies within the range of their sensing radius to move to. 
The exploration search for a new point will start after the cant decides if it will stay on the same lag-level or jump to a higher level. If it is moving on the same level, the cant will generate a random number between $[0,1]$ as an angle bisector, or between $[-1,1]$ if jumping to a higher level. An angle is then calculated using the bisector:  $\theta=angle\_bisect \times \pi$, which will be the direction of the cant's next coordinate move using these formulae: 
\[
\centering
    x_{new} \gets x_{old} + \rho * cos(\theta); \quad\quad  y_{new} \gets y_{old} + \rho * sin(\theta)
\]    
% Once a cant decides if it is climbing or staying within the same lag-level, it will generate an angle bisector that is either a random number between $[0, 1]$ if the current and next point are on the same lag-level or a number in $[-1, 1]$ if the current and next points are on different lag-levels. 
% This angle bisector is used to calculate the angle of the next movement of the cant: $\theta=angle\_bisect \times \pi$. 
% The movement angle is then subsequently used to calculate the next $x$ and $y$ coordinates of the next position of the cant: $x_{new} \gets x_{old} + \rho * cos(\theta)$, $y_{new} \gets y_{old} + \rho * sin(\theta)$. These points are also saved for potential future pheromone modification. 

\paragraph{Cants Evolution:} 
Individual CANTS agents, cants, are born/initialized with uniformly distributed random exploration/exploitation and sensing-range coefficients. In addition, two more parameters are introduced to control the speed at which a cant will move (for one step) by controlling the size of the sensing radius using this $2^{nd}$ degree polynomial formula: 
\[
sense\_range_{precieved} =  max \Big( sense\_radius - \big( y^2 \times r_1 + y \times r_2 \big), 0.1 \Big)
\]
\begin{wrapfigure}[17]{r}{0.6\textwidth}
    \centering
    \vspace{-15pt}
    \includegraphics[width=.99\textwidth, height=.55\textwidth]{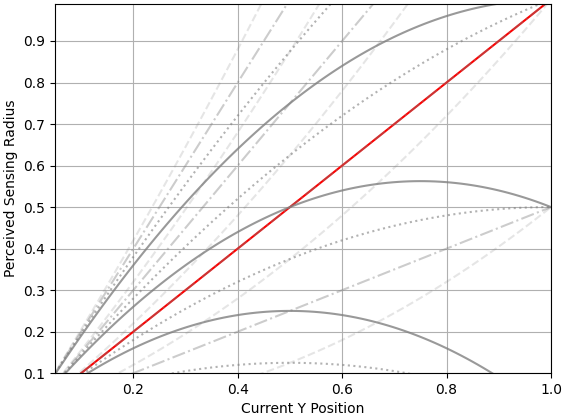}
    \caption{\centering \textit{Potential cant speed pattern based on its $y$ position, $r_1$, and $r_2$ values.}}
    \label{fig:cant_speed}
\end{wrapfigure}
where $r_1$ and $r_2$ parameters control the size of the single cant's perceived sensing range based on how far the cant is from the input node(s) ($y$ spacial distance from inputs). The bigger the sensing range, the more the cant is able to conceptually move longer steps at a time, and the smaller the sensing rage, the smaller the steps the cant can do, and yet, the lowest range value is set at $0.1$ to guarantee that the agent will not completely loose its pheromone sense and halt in position. 
The polynomial aspect of the formula used will eventually control the pattern of the cant movement speed: {\it a)} slower at the begging and faster as it approaches the outputs, or {\it b)} slower at the begging, speeds up in the middle, and slows as it approaches the outputs. Figure~\ref{fig:cant_speed} illustrates those patterns.

% \begin{wrapfigure}[]{r}{0.6\textwidth}
% % \begin{figure}
%     \centering
%     \includegraphics[width=.98\textwidth, height=.5\textwidth]{img/ants_hpc.pdf}
%     \caption{\textit{The CANTS asynchronous design.}}
%     \label{fig:async_ants}
% % \end{figure}
% \vspace{-0.2cm}
% \end{wrapfigure}  

% \begin{figure}
%     \centering
%     \includegraphics[width=.98\textwidth]{img/region_moves.png}
%     \caption{"BLA BLA"}
%     \label{fig:cant_speed}
% \end{figure}

Although these cants parameters are uniquely generated for the individual cants, they evolve between the CANTS iterations using a genetic algorithm that evolves those parameters through mutation and cross-over operations, based on the best performing generated NN (see $cant.Evolve()$ in Algorithm~\ref{alg:cants_pseudo}). 

Incorporating the above behavioral traits provides more flexibility to the optimization agents themselves and devising this evolution mechanism makes those parameters self-tune during evolution instead of being treated as extra hyperparameters.

\paragraph{Condensing Cant Paths to RNN Nodes:} 
The points on different cants paths can be in a spacial proximity that can be also mapped to neural proximity, thus, represent a type of redundancy. To avoid having extraneous neurons in the neural space, the points on the cants' paths are clustered using the DBSCAN algorithm~\cite{ester1996density}, condensing the resulting clusters to centriods. 
% After cants choose the points in their paths from the inputs to the outputs, the points in the search space are clustered using the DBSCAN algorithm~\cite{ester1996density} to condense the points to centroids. 
After that, the points of neighborhood inside a given cluster are represented by the centriod of the cluster, and the centriod is translated to a neuron in the generated neural architecture (see Figures~\ref{fig:ants_multi_path} and~\ref{fig:ants_condensed_path}).
% The points of the segments of the cants' paths are then shifted to the relevant centroids that they belong to in the search space and the resulting new points become the nodes of the generated RNN architecture (see Figures~\ref{fig:ants_multi_path} and~\ref{fig:ants_condensed_path}). 
The node types are picked by a pheromone-based discrete local search, as is done in the discrete space ANTS algorithm. Each of these node types at the selected point will have their own pheromone values that drive probabilistic selection.

\paragraph{Communal Weight Sharing:} 
To avoid retraining the neural parameters of newly-generated RNN from scratch, BP-Cants uses a communal-weight-sharing strategy is deployed to offer an advanced starting point to the generated RNNs, exploiting parametric values of previously trained RNNs. (refer to~\cite{elsaid2021continuous} for more details).
% The centroid points (\ie, the RNN node points in the search space) in CANTS retain the weights of all the out-going edges from those nodes. Each newly-created centroid is assigned a weight value which is passed to the edges of the generated RNN. In the case, when a centroid does not have any previously created centroid within its cluster/range, randomly initialized weights are assigned to those outgoing edges either uniformly at random between $-0.5$ and $0.5$, or via the Kaiming~\cite{he2015delving}~or Xavier~\cite{glorot2010understanding}~initialization strategies. If there were previously-created centroids in the clustering region, the weight values assigned to the generated RNN nodes are the average of the weights of those existing centroids. The weights of points in a centroid cluster are then updated after an RNN is trained by calculating the averages of the original centroid-cluster point weight values and all the weights of the outgoing edges of the corresponding node (after training). The updated weights can then be used to initialize new centroid weights when they lie within their cluster after DBSCAN is applied in the following iteration. 
In BP-Free CANTS, the generated RNNs are only tested without BP training, thus, the communal weight sharing is carried throughout the optimization process without requiring any updates from the generated RNNs.

\paragraph{Pheromone Deposit:}
When a candidate RNN wins a place in the population, the manager process increments the pheromone values of the architecture's corresponding space centroids by a constant value as a reward.
% For each successful candidate RNN, \ie, an RNN that performs at least better than the current worst one in the population, the corresponding centroids for its RNN nodes in the search space are rewarded by increasing their pheromone values by a constant value.
Since they contribute to the creation of centroids by their pheromone values, the points in proximity to centroids, i.e. the cluster of points used to generate the centroid, are also rewarded by increasing their pheromone levels by a fraction of the same constant value, depending on their distance from the centroid.
The space points' pheromone values is capped at a maximum threshold to avoid having  points of super-attraction to cants, causing the optimization to prematurely conclude. 
% The values of the pheromones have a maximum limit to avoid generating points that are overly attractive to the cants, which could result in premature convergence. 
% The other points in the centroid-clusters (other than the centroids) are also rewarded by fractions of the same constant value, depending on how far the point is from the centroid of the cluster.

\paragraph{Pheromone Volatility:}
Pheromones regularly decay through the process iterations, regardless of the RNNs' performance reported by workers. The decay occurs by a constant value and, after a predetermined minimum threshold, the point vanishes from the search space. Wiping out points with faint pheromone traces allows the search space to git rid of pheromones tiny residual that might impede cant-to-cant communication as well as dragging the overall optimization process. This pheromone degradation also prevents the algorithm from getting prematurely stuck in local minima.

{
% \color{red}
        \paragraph{Time Complexity of the Algorithms:} While the time complexity of the various ANTS algorithms is important, it should be noted that the time spent generating recurrent neural networks with these algorithms is orders of magnitude less than the time to evaluate the fitness of the generated neural networks (which requires training and validation for ANTS and BP-CANTS, and only validation for BP-Free CANTS). This is highlighted in Figure~\ref{fig:time_accum}.\\
        For ANTS and BP-CANTS, each generated recurrent neural network (which may have tens of thousands of trainable parameters) needs to be trained using backpropagation through time for a number of epochs. As all the ANTS algorithms use an asynchronous distributed/parallel execution strategy, training potentially hundreds of RNNs simultaneously on distributed compute nodes while the master process which generates new RNNs spends most of its time waiting for more work requests from the workers (it may take fractions of a second to generate a network, however, training them and then evaluating them on the validation dataset can take multiple minutes, even for few number of epochs). For BP-Free CANTS, the forward pass of the network over the validation data to evaluate the network without backpropagation is still orders of magnitude faster than the time for the algorithm to generate the networks.\\
        The time complexity of BP-Free CANTS is O($P$ log $P$), where $P$ is the number of pheromone points in the search space. This is due to the complexity being bound by the DBSCAN operation on the pheromones in the search space (all other operations, e.g., network construction are linear based on the clusters generated by DBSCAN). New pheromone values are added every time cant agents move through the search space. These pheromone values degrade over time, and are removed when they fall below a given threshold. Unfortunately, due to the stochastic nature of the algorithm it is not possible to compute an upper bound on the maximum number of pheromones possible. That being said, given our experimental results presented in Figure~\ref{fig:time_accum} and discussed in Section~\ref{sec:time_benchmarking}, RNN generation time did not appear to grow in an unbounded manner in our experiments. A potential area of future work (if this proves to be a computational bottleneck) would be to add in strict limits to the possible number of pheromones present in the search space at any time, by for example removing the oldest pheromones or combining clustered pheromone traces.

}
\section{Results}
\label{sec:results}

This work compares BP-Free CANTS to the previous state-of-the-art ANTS and BP-CANTS algorithms on three real world datasets related to power systems. All three methods were used to perform time series data prediction for a parameter, which have been used as benchmarks in prior work. 
% Main flame intensity was used as the prediction parameter from the coal plant's burner, 
For the coal plant data, note that net plant heat rate was used from the coal plant's boiler. Experiments were also performed to investigate the effect the number of cants. 
% , and Average power output was used from the wind turbines. Experiments were also performed to investigate the effect of CANTS hyper-parameters: the number of cants and cant sensing radii, $\epsilon$. 
%\zimeng{maybe put the sentences about dataset into dataset section}

  \paragraph{\textbf{Computing Environment}}
\label{sec:computing}
The results for ANTS, BP-CANTS), and BP-Free CANTS were obtained by scheduling the experiment on Rochester Institute of Technology's high performance computing cluster with $64$ Intel\textsuperscript{\textregistered} Xeon\textsuperscript{\textregistered} Gold $6150$ CPUs, each with $36$ cores and $375$ GB RAM (total $2304$ cores and $24$ TB of RAM). The experiments used $3$ nodes ($108$ cores) and also took approximately $4$ weeks to complete the experiments.

\paragraph{\textbf{Datasets}}
\label{sec:datasets} 
The dataset used, which is derived from coal-fired power plant sensor readings, has been previously made publicly available on the EXAMM repository to encourage further study in time series data prediction and reproducibility~\footnote{\href{https://github.com/travisdesell/exact/tree/master/datasets/}{https://github.com/travisdesell/exact/tree/master/datasets/}}. The dataset comes from measurements collected from $12$ burners of a coal-fired power plant. 

The dataset is multivariate and non-seasonal, with $12$ input variables (potentially dependent). These time series are very long, with the burner data separated into $7000$ time step chunks. The dataset is sampled and  separated into a training set of $1875$ steps and a test set of $625$ steps (per minute recordings).

% \begin{figure}
%     \RawFloats
%     \centering
%     \begin{minipage}{0.49\textwidth}
%         \centering
%         \includegraphics[width=.99\textwidth, height=0.3\textheight]{./img/boxplt_mae_cants_num_ants.pdf}
%         \caption{\label{fig:cants_num_ants}  {CANTS w/ varying \# of agents.}}
%     \end{minipage}
%     \hfill
%     \begin{minipage}{0.49\textwidth}
%         \centering
%         \includegraphics[width=.99\textwidth, height=0.3\textheight]{./img/boxplt_mae_cants_sensing_range.pdf}
%         \caption{\label{fig:cants_radii} {CANTS w/ different sensing radii.}}
%     \end{minipage}
% \end{figure}

\subsection{On the Number of Cant Agents}
An experiment was conducted to determine the effect that the number of cant agents would have on the performance of the BP-Free CANTS algorithm. The experiment focused on the net plant heat rate feature from the coal-fired power plant boilers dataset. The number of ants that were simulated and evaluated were $5$, $10$, $15$, $25$, $35$, and $50$. 
% The results, shown in Figure~\ref{fig:fit_scatter}, show that, as the number of cants are increased, ANTS performance fluctuated, but the trend of the min/max values shows a decline. Both versions of CANTS performance steadily increased till 25 cants and then saturated at the rest of number of cants. This shows that the number of cant agents is an important hyper-parameter and requires tuning, potentially exhibiting ``sweet spots'' that, if uncovered, provide strong results. 

% \begin{figure}
%     \centering
%     \includegraphics[width=.9\textwidth]{img/fit_scatter.pdf}
%     \caption{\bplesscants Performance Compared to CANTS w/ BP \& ANTS}
%     \label{fig:fit_scatter}
% \end{figure}

% \begin{figure}
%     \centering
%     \includegraphics[width=.9\textwidth]{img/fit_box.pdf}
%     \caption{\bplesscants Performance Compared to CANTS \& ANTS}
%     \label{fig:fit_box}
% \end{figure}

\subsection{Algorithm Benchmark Comparisons}
To compare the three different NAS strategies, each experiment was repeated $10$ times (trials) to facilitate a statistical comparison and the algorithms were set to the following amounts of computational processing: 
the ANTS and BP-CANTS (the memetic algorithms) were simulated over $1000$ total steps with $30$ epochs of local backpropagation applied (to tune candidate RNNs), whereas for the proposed non-memetic BP-Free CANTS, $3000$ steps were simulated without any backpropagation. 
%generate $3000$ RNNs per trial. % for BP-CANTS, $1000$ RNNs per trial for BP-Free CANTS and ANTS. 
\bplesscants was given more optimization iterations because it is faster to finish (as will be discussed later in the section), while BP-CANTS and ANTS consume more time since they perform BP epochs per round/iterations of their respective optimization process. For both BP and BP-Free CANTS, the sensing radii of the cant agents and exploration instinct values were generated uniformly via $\sim U(0.01, 0.98)$ when the cants were created/initialized, initial pheromone values were set to $1$ and the maximum was kept at $10$ with a pheromone decay rate set to $0.05$. 
For the DBSCAN module, clustering distance was set to $0.05$ with a minimum point value of $2$. 
The population size used was $10$. ANTS, BP-CANTS, and BP-Free CANTS had a maximum recurrent depth of $5$ and the predictions were made over a forecasting horizon of $1$. 
The generated RNNs were each allowed $30$ epochs of back-propagation for local search/fine-tuning for the memetic algorithms ANTS and BP-CANTS. ANTS utilized the hyper-parameters previously reported to yield best results~\cite{ororbia2019examm,elsaid2020ant}.

% \begin{figure}
%     \centering
%     \includegraphics[width=.9\textwidth]{img/time_scatter.pdf}
%     \caption{Time Consumed By \bplesscants, CANTS, \& ANTS}
%     \label{fig:time_scatter}
% \end{figure}

% \begin{figure}
%     \centering
%     \includegraphics[width=.9\textwidth]{img/time_box.pdf}
%     \caption{Time Consumed By \bplesscants, CANTS, \& ANTS}
%     \label{fig:time_box}
% \end{figure}

\begin{figure}[!t]
    \centering
    \includegraphics[width=.8\textwidth]{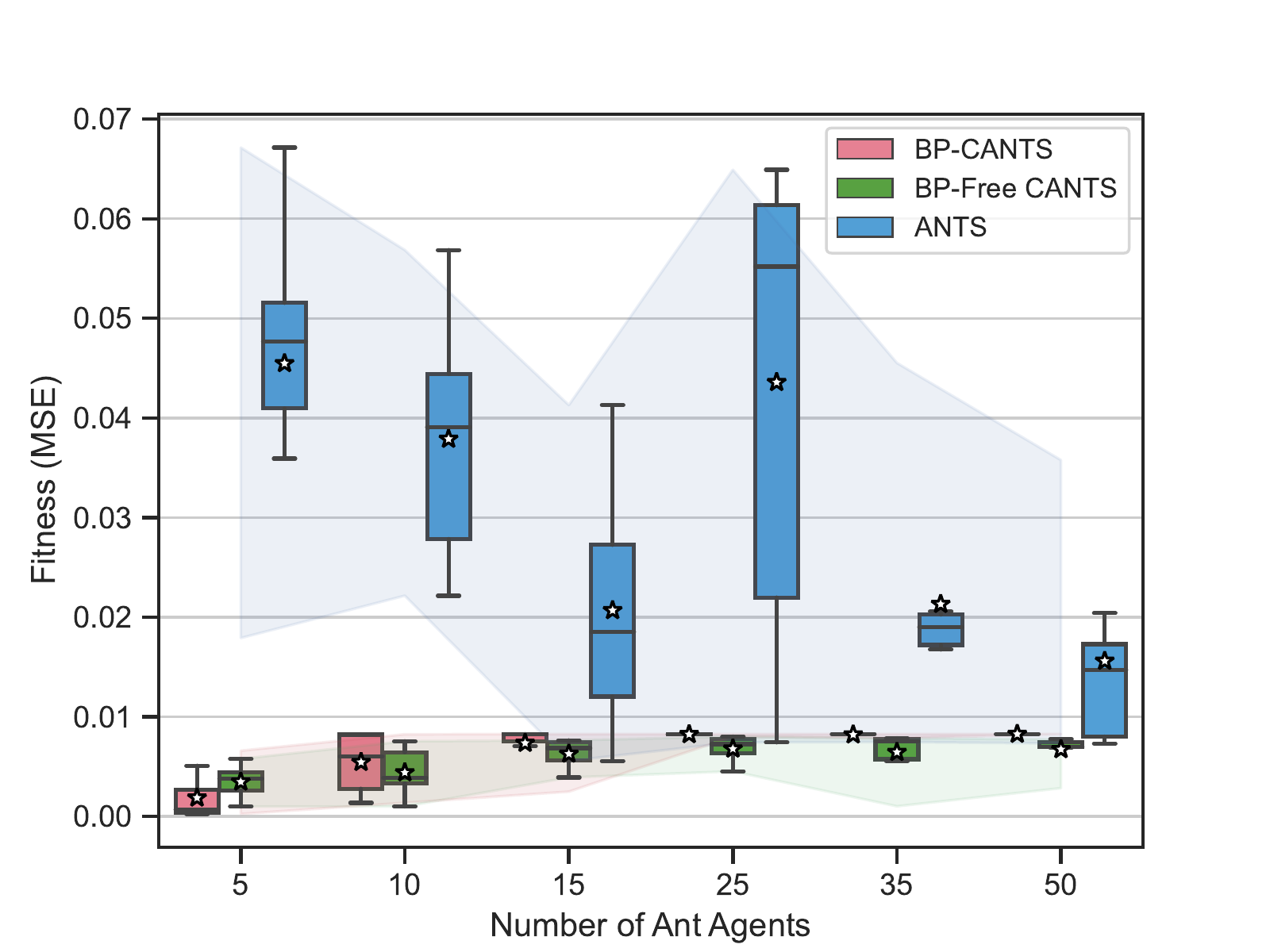}
    \caption{\textbf{\textit{ \bplesscants Performance (as a function of the number of agents/particles) of BP-Free CANTS compared to CANTS and ANTS. Note that a lower value of the fitness/loss is better (given that optimization is focused on minimization).}}}
    \label{fig:fit_box_scatter}
\end{figure}

\begin{table}[] 
\caption{\label{tab:fit} \textbf{\textit{Average Fitness}}}
\begin{tabular}{c|l|l|l} 
        & \multicolumn{3}{c}{\large\textbf{Average Fitness (MSE)}} \\
\textbf{No. Agents} & \textbf{BP-Free CANTS} & \textbf{BP CANTS} & \textbf{ANTS} \\
\midrule
\textbf{ 5} & 0.0035 & \colorbox{lightgray}{0.0018} & 0.0455	\\
\textbf{10} & \colorbox{lightgray}{0.0044} & 0.0058 & 0.0379	\\
\textbf{15} & \colorbox{lightgray}{0.0063} & 0.0073 & 0.0207	\\
\textbf{25} & \colorbox{lightgray}{0.0068} & 0.0082 & 0.0436	\\
\textbf{35} & \colorbox{lightgray}{0.0064} & 0.0082 & 0.0216	\\
\textbf{50} & \colorbox{lightgray}{0.0067} & 0.0083 & 0.0156	\\
\end{tabular}
\end{table}

\subsubsection{Performance Benchmarking}
The results shown in Figure~\ref{fig:fit_box_scatter} -~which compare the performance of \\ BP-Free CANTS, BP-CANTS, and ANTS as described in the experimental settings above~- are measurements of the range of mean squared error (MSE) of each algorithm's best-found RNNs. The figure uses a box-plot to illustrate the minimum, maximum, median (line in interquartile range), and average (star markers) of the different trails of the experiments, without showing the outliers. The shadow-shapes bound the area between the maximum and the minimum values of the fitness within the different  trails of experiments (but with the outliers counted). {
% \color{red}
Table~\ref{tab:fit} depicts the average MSE for the 3 methods, showing that BP-Free CANTS is at the lead in the vast majority of the experiments.}

Desirably, both BP-CANTS and the proposed BP-Free CANTS - outperformed ANTS in all of the experiment trails. The mean and the average of the ANTS results over the different number of ants/agents seemed to consistently improve from $5$ ants to $15$ ants, but then some less accurate results started to show up/appear when $25$ ants were used. Nevertheless, the ANTS results continued to improve as the trend continued past $35$ and $50$  ants. Generally, the ANTS algorithm's performance exhibits improved MSE as more agents were used. 

While BP-CANTS slightly outperformed \bplesscants with $5$ cants, the latter exhibited consistently better performance than the former for all other numbers of agents. The saturation of the performance curve of CANTS against the number of cants indicates that for the dataset used, the optimum number of optimization agents was with the fewest number of cants $5$, however BP-Free CANTS was finding still finding comparative best newtworks at $35$ and $50$ cants. Note that BP-Free CANTS obtained these good-quality results (with respect to fitness/loss) using a higher number of optimization iterations but crucially at a significantly lower computational time cost (see next section). 
%The fluctuation and degradation in ANTS performance is most likely the result of a lower number of iterations compared to previous experiments~\cite{elsaid2021continuous}. 

\begin{figure}[!t]
    \centering
    \includegraphics[width=.8\textwidth]{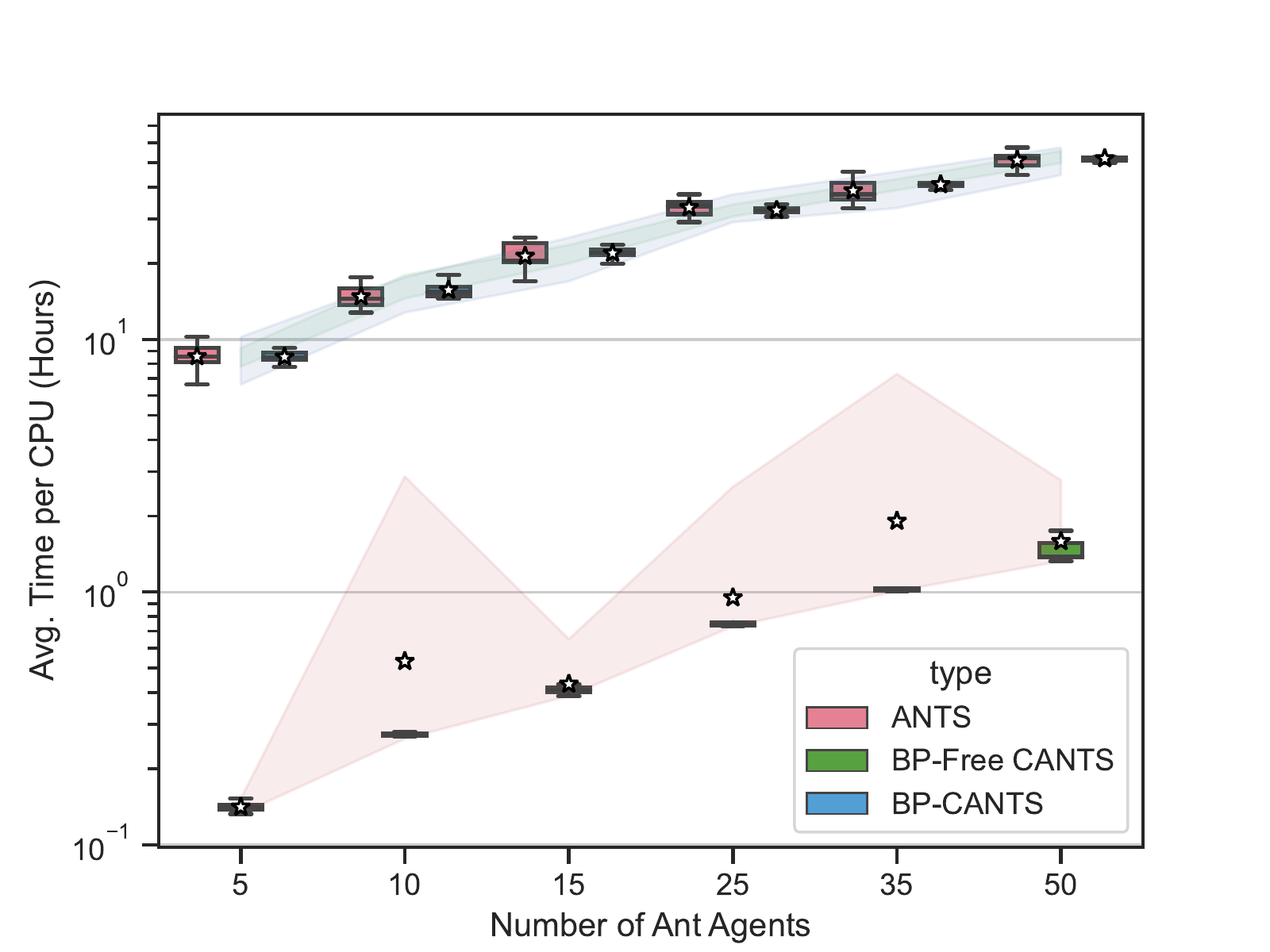}
    \caption{\textbf{\textit{Time consumed by the \bplesscants, CANTS, and ANTS algorithms (lower/closer to zero is better). Note that the y-axis is a logarithmic scale, which shows BP-Free CANTS operating orders of magnitude faster than BP-CANTS and ANTS.}}}
    \label{fig:time_box_scatter}
\end{figure}

\subsubsection{Time Benchmarking}
\label{sec:time_benchmarking}
Figure~\ref{fig:time_box_scatter} illustrates the average time consumed by a CPU in the experiments done for each optimization method (ANTS, BP-CANTS, BP-Free CANTS) across the different trails of the experiments. As observed in the figure, there is a significant gap between the time periods taken by BP-Free CANTS and both ANTS and BP-CANTS. Not only was the performance of BP-Free CANTS significantly better than CANTS for most of the used agent counts, even with the increased evolution operations undertaken by BP-CANTS to evolve its agents during the simulated optimization iterations, it still finished orders of magnitude faster than ANTS and BP-CANTS because it did not have to apply BP during its iterations.  The total operation time of CANTS (with BP) and ANTS were notably similar to each other, however this is to be expected given both were allowed to perform the same number of optimization iterations and BP epochs. {
% \color{red}
Table~\ref{tab:time} depicts the large gap between the BP-Free CANTS average consumed time at different number of agents utilization, and the BP-CANTS and ANTS, which both use backprobagation in their optimization process. On average, BP-Free CANTS is faster than and BP-CANTS and ANTS by $98.5\%$, and $96.1\%$ respectively.}

\begin{figure}[h]
    \centering
    \includegraphics[width=.9 \textwidth]{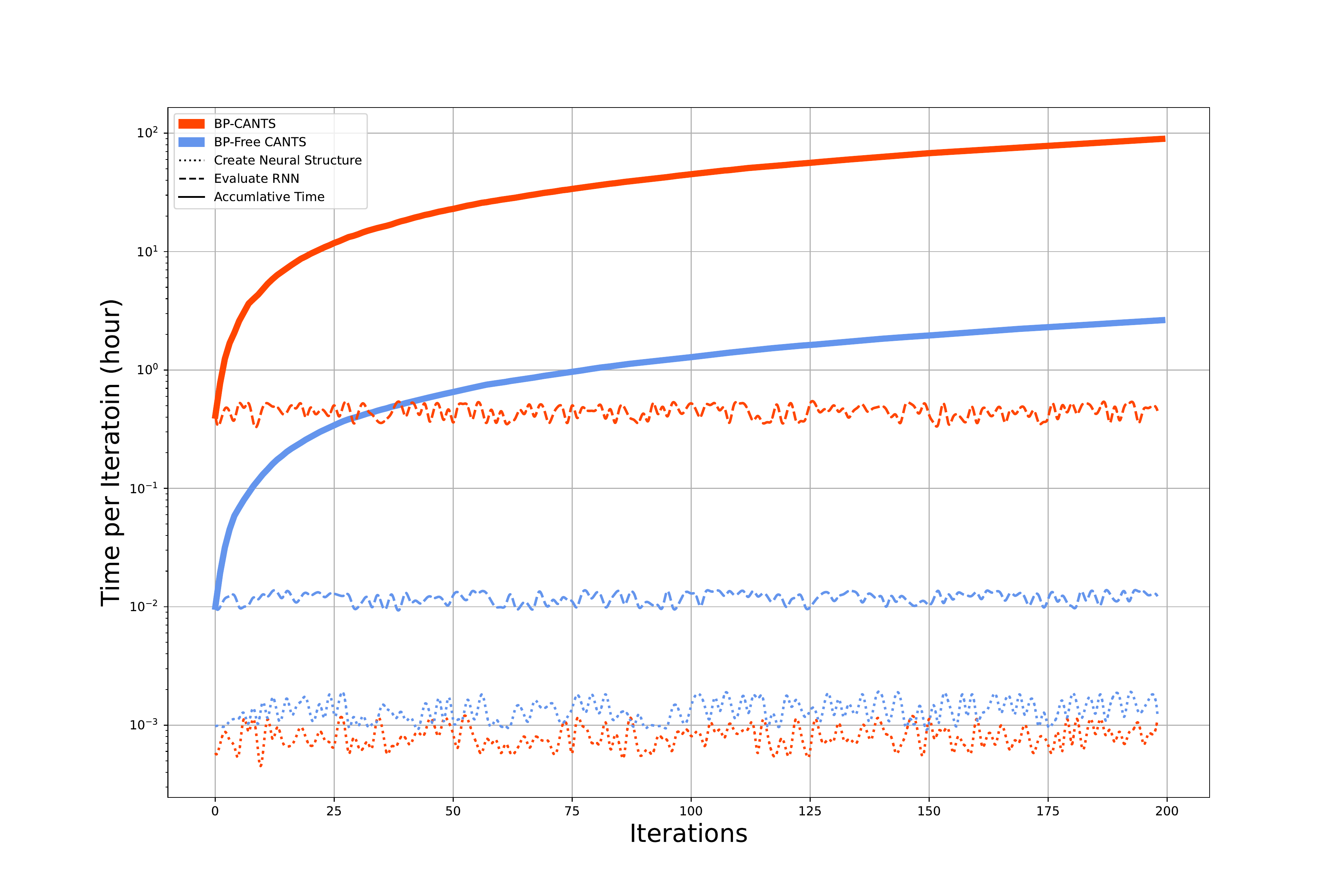}
    \caption{\label{fig:time_accum}\textit{\textbf{Time Consumed by BP-Free CANTS vs. Time Consumed by BP-CANTS:} Time consumed to generate neural structures (dotted lines) is about the same for both methods. BP-CANTS consumes more evaluation time (dashed lines) because they use BP. Adding the accumulative generation and evaluation time gives the solid curves for both BP-CANTS and BP-Free CANTS.
     }}
\end{figure}

{
% \color{red}
Figure~\ref{fig:time_accum} illustrates the time consumption-difference of BP-Free CANTS and BP-CANTS. The two methods were allowed to generate $200$ neural structures using $8$ CPUs and $35$ agents. The figure shows the time consumed to generate the neural structure, where the two methods timing were very close with BP-Free CANTS taking a bit more time to process due to the extra $4^{th}$ dimension. Observably, the evaluation time of BP-CANTS was an order of magnitude higher than that of the BP-Free CANTS because the former used BP, while the latter does not. The figure also shows the accumulative time of neural-structure-generation and the RNN-evaluation of the two methods as the iterations progress. The great difference between the accumulative time of BP-CANTS and BP-Free originates from the time consumed by BP. 
The slope of the cumulative time is higher at early iterations compared to later ones because the search space is fairly empty (no pheromone deposits) at the beginning of the iterations, but as iterations increase, the search space is gradually filled, causing the agents to create more neurons, which increases the evaluation time. The curves then saturates a bit because of the pheromone-evaporation effect, which maintains the pheromone traces at about constant level.
}

\begin{table}[] 
\caption{\label{tab:time} \textbf{\textit{Average Time Consumed}}}
\begin{tabular}{c|l|l|l} 
& \multicolumn{3}{c}{\large\textbf{{Average Time (hrs)}}} \\
% \cmidrule{2-4}
\textbf{No. Agents} & \textbf{BP-Free CANTS} & \textbf{BP CANTS} & \textbf{ANTS} \\
\midrule
\textbf{5}  & \colorbox{lightgray}{0.14} & 7.75 & 7.70	\\
\textbf{10} & \colorbox{lightgray}{0.53} & 14.24 & 13.29	\\
\textbf{15} & \colorbox{lightgray}{0.43} & 19.94 & 19.21	\\
\textbf{25} & \colorbox{lightgray}{0.95} & 23.57 & 26.58	\\
\textbf{35} & \colorbox{lightgray}{1.91} & 26.08 & 27.13	\\
\textbf{50} & \colorbox{lightgray}{1.43} & 33.08 & 35.82	\\
\end{tabular}
\end{table}

\begin{comment}
\begin{figure}
    \centering
    \includegraphics[width=.99\textwidth]{./img/rnn_01_1.pdf}
    \caption{{Resulting Structure Iteration 1}}
    \label{fig:cants_rnn_1}
\end{figure}

\begin{figure}
    \centering
    \includegraphics[width=.99\textwidth]{./img/rnn_02_10.pdf}
    \caption{{Resulting Structure Iteration 10}}
    \label{fig:cants_rnn_2}
\end{figure}

\begin{figure}
    \centering
    \includegraphics[width=.99\textwidth]{./img/rnn_03_11.pdf}
    \caption{{Resulting Structure Iteration 11}}
    \label{fig:cants_rnn_3}
\end{figure}

\begin{figure}
    \centering
    \includegraphics[width=.99\textwidth]{./img/rnn_04_108.pdf}
    \caption{{Resulting Structure Iteration 108}}
    \label{fig:cants_rnn_4}
\end{figure}

\begin{figure}
    \centering
    \includegraphics[width=.99\textwidth]{./img/rnn_05_364.pdf}
    \caption{{Resulting Structure Iteration 364}}
    \label{fig:cants_rnn_5}
\end{figure}

\begin{figure}
    \centering
    \includegraphics[width=.99\textwidth]{./img/rnn_06_544.pdf}
    \caption{{Resulting Structure Iteration 544}}
    \label{fig:cants_rnn_6}
\end{figure}

\begin{figure}
    \centering
    \includegraphics[width=.99\textwidth]{./img/rnn_07_995.pdf}
    \caption{{Resulting Structure Iteration 995}}
    \label{fig:cants_rnn_7}
\end{figure}
\end{comment}

\section{Discussion and Future Work}
\label{sec:conclusion}

This work proposes the backpropagation-free continuous ant-based neural topology search (BP-Free CANTS), a novel, nature-inspired, and non-memetic optimization method that employs a 4D continuous space to conduct unbounded neural architecture search (NAS). CANTS provides a unique strategy to overcome the key limitations of constructive neuro-evolutionary strategies (which are often prematurely trapped at local minimas) as well as other NAS strategies bonded by search space limits . Moreover, BP-Free CANTS expands over previous ACO-centric approaches by crucially mapping the search space of the synaptic weights of the evolved neural networks as a fourth dimension in a 4D space, unifying both the search for optimal neural structure and synaptic weight values. 

Experimental evaluations were carried on BP-Free CANTS to validate its performance in automatically designing recurrent neural networks (RNNs) used in time-series predictions, using a real-world dataset in the power systems domain. We compared our procedure to a state-of-the-art meta-heuristic optimization approach, ANTS (a discrete space ant colony NAS algorithm) as well as the powerful memetic/backprop-centric algorithm CANTS (BP-CANTS). Our results demonstrate that BP-Free CANTS improves over ANTS and BP-CANTS while running substantially and significantly faster. %while also being simpler to use and tune, only requiring $8$ hyper-parameters as opposed to the $16$ hyper-parameters of the other strategy.

This study presents important steps in generalizing ant colony optimization algorithms to complex, continuous search spaces, specifically for unbounded graph optimization problems (with NAS as the key target application), opening up a number of promising avenues for future work. In particular, for BP-Free CANTS,  while the search space is continuous in each two-dimensional plane (or time step) of our temporal stack, there is still the (maximum) number of discrete levels that a user must specify. Therefore, a promising extension of our algorithm would be to make the search space continuous across all three dimensions, removing this parameter entirely, allowing pheromone placements to guide the depth of the recurrent connections. This could have implications for discrete-event, continuous-time RNN models \cite{mozer2017discrete}, which attempt to tackle a broader, more complex set of sequence modeling problems.  Additionally, BP-Free CANTS can provide a well performing solution to the design and training of networks which have activation functions or other components for which derivatives cannot be calculated (which preclude them from utilizing backpropagation), such as models with sampled activities (e.g., Bernoulli distribution) or even discrete ones such as spiking neural networks. 
{
% \color{red}
A potential disadvantage of BP-Free CANTS is that if enough computational resources are available to offer BP-based NAS methods a significant number of training epochs (without worry about cost or emissions), the accuracy of the resulting models will be higher than of BP-Free Cants. Nevertheless, the computational cost of BP is always a burden in NAS methods, and a good reason to seek faster alternatives"~\cite{miller1990cmac,1007668,406684,ORKCU20113703}.
}
Finally, and potentially the most interesting, ACO algorithms including the one designed in this work, generally focus on utilizing only one single colony. According to  myrmecologists\footnote{Myrmecology: The branch of entomology focusing on the scientific study of ants.}, it would prove fruitful to view and design synthetic ant colonies as living organisms themselves, with ants as their living ``cells''~\cite{gordon2010ant}, potentially offering a flexible, scalable simulation framework for modeling how ant agents might achieve more complex tasks related to overall survival (i.e., a colony of ant colonies). Expanding this algorithm to other domains, such as the automated design of convolutional neural networks (for computer vision) or to other types of recurrent temporal networks, such as those used for natural language processing, would further demonstrate the  broad applicability of this nature-inspired approach.

\section{Acknowledgements}
The experiments carried out in this work were facilitated by the computational resources and support of: 

\begin{itemize}
    \item[--]The NSF Advanced Cyberinfrastructure Coordination Ecosystem: Services \& Support (ACCESS - formally XSEDE)\footnote{Grant number: CIS220035}.
    \item[--] The Research Computing at the Rochester Institute
       of Technology\footnote{Rochester Institute of Technology. https://doi.org/10.34788/0S3G-QD15}. 
\end{itemize}

\bibliographystyle{acm}
\bibliography{bibliography_merged}  %%% Uncomment this line and comment out the ``thebibliography'' section below to use the external .bib file (using bibtex) .





% \end{thebibliography}

\end{document}